\documentclass[nai,crcready]{iosart2x}

\usepackage{soul}
\usepackage[utf8]{inputenc}
\usepackage[small]{caption}
\usepackage{dsfont}
\usepackage{bbold}
\usepackage{graphicx}
\usepackage{url}
\usepackage{prettyref}
\usepackage{mathptmx}
\usepackage{amsmath,amssymb}
\usepackage[ruled,vlined]{algorithm2e}

\usepackage{caption}
\usepackage[subrefformat=parens]{subcaption}
\usepackage{booktabs}

\newcommand{\mbf}[1]{ {\mathbf #1} }

\newcommand{\rmin}{\mbox{min}}
\newcommand{\rmax}{\mbox{max}}

\newcommand{\riff}{\mbox{iff}}
\newcommand{\rpdg}{\mbox{pdg}}

\newcommand{\matr}[1]{\mathbf{#1}}

\newtheorem{proposition}{Proposition}%
\newtheorem{definition}{Definition}
\newtheorem{example}{Example}

\pubyear{0000}
\volume{0}
\firstpage{1}
\lastpage{1}

\begin{document}

\begin{frontmatter}

%\pretitle{}
\title{Towards end-to-end ASP computation} %no use of "end-to-end ASP computation" by Google
\runtitle{Towards end-to-end ASP computation}
%\subtitle{}

% For one author:
%\author{\inits{N.}\fnms{Name1} \snm{Surname1}\ead[label=e1]{first@somewhere.com}}
%\address{Department first, \orgname{University or Company name},
%Abbreviate US states, \cny{Country}\printead[presep={\\}]{e1}}

% Two or more authors:
\begin{aug}
\author[A]{\inits{T.}\fnms{Taisuke} \snm{Sato}\ead[label=e1]{taisukest@gmail.com}%
%\thanks{Corresponding author. \printead{e1}.}
}
\author[B]{\inits{A.}\fnms{Akihiro} \snm{Takemura}\ead[label=e2]{atakemura@nii.ac.jp}}
\author[C]{\inits{K.}\fnms{Katsumi} \snm{Inoue}\ead[label=e3]{inoue@nii.ac.jp}}
%\author[B]{\inits{N.-N.}\fnms{Name3-Name3} \snm{Surname3}\ead[label=e3]{third@somewhere.com}}
%\address[A]{Department first, \orgname{National Institute of Informatics},
\address[A]{\orgname{National Institute of Informatics}, \cny{Japan}\printead[presep={\\}]{e1}}
%\address[B]{Department first, \orgname{University or Company name},
\address[B]{\orgname{National Institute of Informatics},
%\cny{Japan}\printead[presep={\\}]{e2,e3}}
\cny{Japan}\printead[presep={\\}]{e2}}
\address[C]{\orgname{National Institute of Informatics},
%\cny{Japan}\printead[presep={\\}]{e2,e3}}
\cny{Japan}\printead[presep={\\}]{e3}}
\end{aug}

%\begin{review}{editor}
%\reviewer{\fnms{First} \snm{Editor}\address{\orgname{University or Company name}, \cny{Country}}}
%\reviewer{\fnms{Second} \snm{Editor}\address{\orgname{First University or Company name}, \cny{Country}
%    and \orgname{Second University or Company name}, \cny{Country}}}
%\end{review}
%\begin{review}{solicited}
%\reviewer{\fnms{First} \snm{Solicited reviewer}\address{\orgname{University or Company name}, \cny{Country}}}
%\reviewer{\snm{anonymous reviewer}}
%\end{review}
%\begin{review}{open}
%\reviewer{\fnms{First} \snm{Open Reviewer}\address{\orgname{University or Company name}, \cny{Country}}}
%\end{review}

\begin{abstract}
We propose an end-to-end approach for Answer Set Programming (ASP) and
linear   algebraically   compute   stable  models   satisfying   given
constraints.  The idea is to implement Lin-Zhao's theorem \cite{Lin04}
together  with  constraints directly  in  vector  spaces as  numerical
minimization of a  cost function constructed from  a matricized normal
logic program,  loop formulas  in Lin-Zhao's theorem  and constraints,
thereby  no  use of  symbolic  ASP  or  SAT  solvers involved  in  our
approach.   We also  propose precomputation  that shrinks  the program
size  and  heuristics  for   loop  formulas  to  reduce  computational
difficulty.   We  empirically  test   our  approach  with  programming
examples including the 3-coloring  and Hamiltonian cycle problems. 

\end{abstract}

\begin{keyword}
\kwd{Answer Set Programming}
\kwd{end-to-end ASP}
\kwd{vector space}
\kwd{cost minimization}
\kwd{loop formula}
\kwd{supported model}
\kwd{stable model}
\end{keyword}

\end{frontmatter}

%%%%%%%%%%% The article body starts:
%\section{}\label{s1}
%\subsection{}\label{s1.1}

\section{Introduction}\label{intro}

Computing stable model semantics \cite{Gelfond88} lies at the heart of
Answer Set Programming  (ASP) \cite{Niemela99,Marek99,Lifschitz08} and
there  have been  a  variety  of approaches  proposed  so far.   Early
approaches such  as \textit{smodels} \cite{Niemela97} used  backtracking.  Then
the concept of  loop formula was introduced and approaches  that use a
SAT  solver  to compute  stable  models  based on  Lin-Zhao's  theorem
\cite{Lin04}  were  proposed.   They include  \textit{ASSAT}  \cite{Lin04}  and
\textit{cmodels}   \cite{Lierler05}  for   example.    Later  more   elaborated
approaches   such   as   \textit{clasp}   \cite{Gebser07,Gebser12}   based   on
conflict-driven  no good  learning have  been developed.   While these
symbolic approaches  continue to  predominate in  ASP, there  has been
another   trend   towards   differentiable   methods.    For   example
Differentiable ASP/SAT \cite{Nickles18a} computes  stable models by an
ASP solver that utilizes derivatives of a cost function.
More  recently  NeurASP   \cite{Yang20}  and  SLASH  \cite{Skryagin22,Skryagin23}
combined deep learning and ASP.  In their approaches, deep learning is
not used in an  end-to-end way to compute stable models,  but used as a
component to  compute and  learn probabilities represented  by special
atoms interfacing to  ASP. 
Takemura and Inoue \cite{Takemura2024} proposed a neurosymbolic learning pipeline that leverages differentiable computation of supported models. Although their method does not specifically address stable model computation, it bypasses the need for a symbolic solver and illustrates how differentiable computation facilitates integration with deep learning.
A step towards  end-to-end computation was
taken  by  Aspis et al. \cite{Aspis2020} and Takemura and Inoue \cite{Takemura2022}.
They formulated the computation of  supported models, a super class of
stable models, entirely as  fixedpoint computation  in vector  spaces,
% \footnote{
% %
% Supported models  and stable  models of  a propositional  normal logic
% program coincide  when the  program is tight  (no infinite  call chain
% through positive  goals) \cite{Fages94,Erdem03}.  Also in  the context
% of probabilistic modeling, an  end-to-end sampling of supported models
% was proposed by Sato and Kojima in \cite{Sato19,Sato20}.
%
% }, 
and  obtain
supported models  represented by binary  vectors.
However, there still remains a  gap between computing supported models and
computing stable models.

In this paper,  we propose an end-to-end approach for  ASP and compute
stable models satisfying given constraints in vector spaces.  The idea
is simple; we implement  Lin-Zhao's theorem \cite{Lin04} together with
constraints directly in vector spaces  as a cost minimization problem,
thereby no use  of symbolic ASP or SAT solvers  involved. 
Since our approach is numerical and relies solely on vector and matrix operations, future work could explore the potential benefits of parallel computing technologies such as many-core CPUs and GPUs.

Technically,  Lin-Zhao's theorem  \cite{Lin04}  states  that a  stable
model  of a  ground normal  logic program  coincides with  a supported
model which  satisfies ``loop formulas'' associated  with the program.
Loop formulas are propositional formulas indicating  how to get out of infinite
loops of top-down  rule invocation. We formulate finding  such a model
as root  finding in  a vector  space of  a non-negative  cost function
represented in terms of the matricized program and loop formulas.  The
problem  is  that  in  whatever  approach we  may  take,  symbolic  or
non-symbolic, computing supported models is NP-hard (for example graph
coloring is  solved by  computing supported models)  and there  can be
exponentially  many loop  formulas to  be satisfied \cite{Lifschitz06}. 
We reduce  this
computational  difficulty in  two  ways.  One  is precomputation  that
removes atoms from the search space which are known to be false in any
stable  model  and   yields  a  smaller  program.   The   other  is  to
heuristically choose  loop formulas  to be  satisfied. 
The latter would mean allowing non-stable model computation, and in 
our continuous approach, we modify the cost function to be affected 
only by these chosen loop formulas. The intuition behind this heuristic 
is that the modified cost function would assign higher cost to models 
that do not satisfy the chosen loop formulas, 
thus driving the search process away from them.

Our   end-to-end   computing   framework   differs   from   those   by \cite{Aspis2020}  and  \cite{Takemura2022}   in  that  they  basically compute  supported models  and  the computing  process  itself has  no mechanism such as loop formulas to exclude non-stable models. 
In addition, any propositional normal logic program is acceptable in our framework, since we impose no restrictions on the syntax of programs like the MD condition \cite{Aspis2020} or the SD condition \cite{Takemura2022}.
More importantly, we incorporate the use of constraints, i.e., 
rules with an empty head, which make ASP programming smooth and practical.
% rules like  $\leftarrow a \wedge \neg b$, which make ASP programming smooth and practical.
% The (multiple definitions) MD condition \cite{Aspis2020}:
%   Every atom has at most one rule that has two or more positive literals
%   or two or more negative literals in the rule body

Hence, our contributions include:
\begin{itemize}
\item a  proposal of end-to-end  computing of stable models  in vector
  spaces for propositional normal logic programs
\item augmentation of the above by constraints
\item  introduction   of  precomputation  and  heuristics   to  reduce
  computational difficulty of stable model computation.
\end{itemize}

\noindent
We add  that since our primary  purpose in this paper  is to establish
theoretical feasibility of end-to-end  ASP computing in vector spaces,
programming examples  are small  and implementation is  of preliminary
nature.
Furthermore, the main search algorithm we propose in this paper is incomplete, in the sense that it does not guarantee reaching a global minimum if it exists, nor it cannot conclusively prove that no solution exists.

In  what  follows,  after preliminaries  in  Section~\ref{prelim},  we
formulate  the computation  of supported  models in  vector spaces  in
Section~\ref{support}    and     that    of    stable     models    in
Section~\ref{stable}.    We   then   show  programming   examples   in
Section~\ref{example}  including  ASP   programs  for  the  3-coloring
problem  and  the Hamiltonian cycle  problem.    We  there  compare  performance  of
precomputation  and  loop formula  heuristics.   Section~\ref{related}
contains related work and Section~\ref{conclusion} is the conclusion.

\section{Preliminaries}\label{prelim}
In  this paper,  we mean  by a  program a  propositional normal  logic
program $P$  which is  a finite set  of {\em rule\/}s  of the  form $a
\leftarrow G$ where $a$ is an atom called the \textit{head}, 
$G$ is a conjunction of literals called the \textit{body} of the rule, respectively.
We equate propositional variables with atoms.
A literal is an atom (positive literal) or its negation (negative literal).
The logical connective \(\neg\) in this paper denotes \textit{negation as failure}.
We suppose  $P$ is  written in  a given  set of  atoms ${\cal  A}$ but
usually assume ${\cal A}$ = atom($P$), i.e., the set of atoms occurring
in $P$.
We use $G^{+}$ and \(G^{-}\) to denote the conjunction of positive and negative literals in $G$, respectively.
$G$ may  be empty. The empty  conjunction is always true.   We call $a
\leftarrow G$ {\em  rule\/} for $a$.  
A rule with an empty head is called a \textit{constraint}.
Let $a  \leftarrow G_1,\ldots, a
\leftarrow G_m$ be rules for $a$ in  $P$.  When $m > 0$, put $\riff(a)
= a  \Leftrightarrow G_1\vee\cdots\vee G_m$.  When  $m=0$, i.e., there
is no  rule for  $a$, put  $\riff(a) =  a \Leftrightarrow  \bot$ where
$\bot$ is a special symbol representing the empty disjunction which is
always false.  We call $\riff(a)$  the {\em completed rule\/} for $a$.
The {\em  completion of\/} $P$,  comp($P$), is defined as  comp($P$) =
$\{\riff(a) \mid \mbox{atom  $a$ occurs in $P$}\}$.  For  a finite set
$S$, we denote the number of elements in $S$ by $|S|$. So $|P|$ is the
number of rules in the program $P$.

An \textit{interpretation} (assignment)  $I$ over a set of atoms  ${\cal A}$ is a
mapping which  determines the truth  value of  each atom $a  \in {\cal
  A}$.  Then the  truth value of a formula $F$  is inductively defined
by $I$, and if $F$ becomes true  evaluated by $I$, we say $I$ satisfies
$F$, $F$ is true in $I$, or $I$  is a \textit{model} of $F$ and write $I \models
F$.  This notation is extended to a  set $F = \{ F_1,\ldots,F_u \}$ by
considering $F$  as a  conjunction $F_1 \wedge\cdots\wedge  F_u$.  For
convenience,  we always  equate $I$  with $\{  a \in  {\cal A}  \mid I
\models a \}$, i.e., the set of atoms true in $I$.
When $I$ satisfies all
rules in  the program $P$,  i.e., $I  \models P$, $I$  is said to  be a
\textit{model} of $P$.  If no rule body contains negative literals, $P$ is said
to be  a definite  program.  In  that case, $P$  always has  the least
model (in the sense of set inclusion)  $\{a \in {\cal A} \mid P \vdash
a  \}$, i.e.,  the set  of atoms  provable from  $P$.  

A  model $I$  of
comp($P$) is a {\em supported model\/}  of $P$ \cite{Apt88,Marek92}.
When $P$  is a definite program, there is at least one supported model, 
and its least model is also a supported model.
In general, there can be multiple supported models for both definite and non-definite programs \(P\).
{\em Stable  model\/}s are  a subclass of  supported models.  They are
defined as follows.   Given a program $P$ and a  model $I$, remove all
rules from  $P$ whose body contains  a negative literal false  in $I$,
then  remove all  negative  literals from  the  remaining rules.   The
resulting  program,  $P^{I}$,  is called  the  Gelfond-Lifschitz  (GL)
reduct of  $P$ by $I$ or  just the reduct of  $P$ by $I$.  It  is a
definite program  and has  the least  model.  If  this least  model is
identical  to  $I$, $I$  is  called  a  {\em  stable model\/}  of  $P$
\cite{Gelfond88}.  
$P$  may have zero  or multiple stable  models, as in the case of supported  models. 
Since   the  existence  of  a   stable  model  is
NP-complete  \cite{Marek99}  and  so   is  a  supported  model,  their
computation is expected to be hard.
Supported models  and stable  models of  a propositional  normal logic
program coincide  when the  program is tight  (no infinite  call chain
through positive  goals) \cite{Fages94,Erdem03}.

Let $F = d_1 \vee\cdots\vee d_h$ be a Boolean formula in $n$ variables
(atoms) in disjunctive  normal form (DNF) where each $d_i$  ($1 \leq i
\leq h$)  is a conjunction of  literals and called a  disjunct of $F$.
When $F$  has no disjunct, $F$  is false.  
%F$ is  called {\em full\/} when every $d_i$ is a conjunction of $n$ distinct literals.

A   walk  in   a  directed   graph  is   a  sequence   $v_1\rightarrow
v_2\rightarrow  \cdots\rightarrow  v_u$  ($u   \geq  1$)  of  vertices
representing the corresponding non-zero  sequence of edges $(v_1,v_2),
\ldots, (v_{u-1},v_u)$.  When $v_u = v_1$, it is said to be \textit{closed}.  
A \textit{cycle}  is  a  closed   walk  $v_1\rightarrow   v_2\rightarrow  \cdots
\rightarrow v_{u}\rightarrow v_1$ where  $\{v_1,\ldots,v_u \}$ are all
distinct. 
A \textit{Hamiltonian cycle (HC)} is a cycle which visits every vertex exactly once.
%
% walk: just a seq. of vertices < trail: a seq. of distinct edges
% < path:a trail where all verticies distinct
%
A \textit{path} is a  walk with  no vertex repeated. 
A directed  subgraph is called {\em  strongly connected\/}  if there are  paths from  $v_1$ to
$v_2$ and from $v_2$ to $v_1$  for any pair of distinct vertices $v_1$
and $v_2$. This ``strongly connected'' relation induces an equivalence
relation over the set of vertices  and an induced equivalence class is
called a \textit{strongly connected component (SCC)}.

The {\em positive dependency graph} $\rpdg(P)$  for a program $P$ is a
directed graph where vertices are atoms  occurring in $P$ and there is
an edge $(a,b)$  from atom $a$ to atom $b$  \textit{if and only if (iff)} there
is a rule $a \leftarrow G$ in  $P$ such that $b$ is a positive literal
in $G$.  
$P$ is said to be {\em tight} \cite{Fages94,Erdem03}\footnote{
In \cite{Fages94}, it is called ``positive-order-consistent''.
} when $\rpdg(P)$ is acyclic, i.e., has no cycle.
A {\em  loop\/} $S =  \{a_1,\ldots,a_u\}$ $(u>0)$ in  $P$ is a  set of
atoms where for any pair of atoms  $a_1$ and $a_2$ in $S$ ($a_1 = a_2$
allowed), there is  a path in $\rpdg(P)$ from $a_1$  to $a_2$ and also
from $a_2$ to  $a_1$.
A singleton loop \(S=\{a\}\) is induced by a self-referencing rule of the form \(a \leftarrow a \wedge G\) where \(G\) is possibly empty, i.e., a self-loop \(a \leftarrow a\).
A \textit{support rule} for \(a\) relative to a loop \(S\) 
is a rule \(a \leftarrow G\) such that \(G^{+} \cap S = \emptyset\).
Given a loop \(L=\{a_1,\ldots,a_u\}\) and its external support rules \(\{a_1 \leftarrow G_{11}, \ldots , a_1 \leftarrow G_{1n},  \ldots, a_u \leftarrow G_{u1}, \ldots, a_u \leftarrow G_{un}\}\), the \textit{(conjunctive) loop formula} is the following implication: \((a_1 \wedge \cdots \wedge a_u) \rightarrow (G_{11} \vee \cdots \vee G_{1n} \vee \cdots \vee G_{u1} \vee \cdots \vee G_{un}) \).
% For a finite  set $S$, we denote  the number of
%elements in $S$ by $|S|$.
%\cite{Liu2006}

We denote vectors by bold lower case letters such as $\mbf{a}$
where  $\mbf{a}(i)$  represents  the   $i$-th  element  of  $\mbf{a}$.
Vectors  are column  vectors  by default.   We  use $(\mbf{a}  \bullet
\mbf{b})$  to stand  for the  inner product  (dot product)  of vectors
$\mbf{a}$ and $\mbf{b}$ of the same dimension.  $\| \mbf{a} \|_1 $ and
$\|  \mbf{a} \|_2  $  respectively  denote the  1-norm  and 2-norm  of
$\mbf{a}$ where $\| \mbf{a} \|_1  = \sum |\mbf{a}(i)|$ and $\| \mbf{a}
\|_2 =  \sqrt{ \sum \mbf{a}(i)^2  }$.  We  use $\mbf{1}$ to  denote an
all-ones  vector of  appropriate dimension.   
An interpretation $I$ over  a set
${\cal A} =  \{a_1,\ldots,a_n\}$ of $n$ ordered atoms  is equated with
an $n$-dimensional binary vector $\mbf{s}_{I} \in \{0,1\}^n$ such that
$\mbf{s}_{I}(i) = 1$ if $a_i$ is  true in $I$ and $\mbf{s}_{I}(i) = 0$
otherwise ($1 \leq i \leq n$). 
$\mbf{s}_{I}$ is called the vectorized $I$.

Bold upper case letters  such as  $\matr{A}$ stand  for a  matrix.  We  use
$\matr{A}(i,j)$ to  denote the  $i,j$-th element  of $\matr{A}$,  $\matr{A}(i,:)$ the
$i$-th  row  of $\matr{A}$  and  $\matr{A}(:,j)$  the  $j$-th column  of  $\matr{A}$,
respectively.  We often  consider one dimensional matrices  as (row or
column) vectors. 
$\|\matr{A} \|_F$ denotes the Frobenius norm of $\matr{A}$, i.e., \(\sqrt{\sum_{i,j}\matr{A}(i,j)^2}\).
Let $\matr{A}, \matr{B}  \in \mathbb{R}^{m \times n}$ be $m  \times n$ matrices.
$\matr{A}  \odot \matr{B}$  denotes their  Hadamard product,  i.e., $(\matr{A}  \odot
\matr{B})(i,j) =  \matr{A}(i,j)\matr{B}(i,j)$ for  $i,j (1\leq  i\leq m,  1\leq j\leq
n)$.   $[\matr{A}; \matr{B}]$  designates  the  $2m \times  n$  matrix of  $\matr{A}$
stacked  onto ${B}$.   We  implicitly assume  that  all dimensions  of
vectors and matrices in various expressions are compatible.
We introduce  a piece-wise  linear function $\rmin_1(x)  = \rmin(x,1)$
that returns the  lesser of 1 and $x$ as  an activation function which
is  related   to  the   popular  activation  function   $\mbox{ReLU}(x)  =
\rmax(x,0)$  by $1-  \rmin_1(x)  = \mbox{ReLU}(1-x)$.   $\rmin_1(\matr{A})$
denotes the  result of  component-wise application of  $\rmin_1(x)$ to
matrix $\matr{A}$.
We also introduce  thresholding notation.  Suppose $\theta$  is a real
number  and  $\mbf{a}$  an  $n$-dimensional  vector.   Then  $[\mbf{a}
  \leq\theta]$  denotes  a  binary  vector  obtained  by  thresholding
$\mbf{a}$  at $\theta$  where  for  $i (1\leq  i  \leq n)$,  $[\mbf{a}
  \leq\theta](i)  =  1$  if  $\mbf{a}(i) \leq  \theta$  and  $[\mbf{a}
  \leq\theta](i) =  0$ otherwise.   $[\mbf{a} \geq\theta]$  is treated
similarly.   We extend  thresholding to  matrices. Thus  $[\matr{A} \leq  1]$
means a matrix such  that $[\matr{A} \leq 1](i,j) = 1$  if $\matr{A}(i,j)\leq 1$ and
$[\matr{A} \leq 1](i,j) = 0$ otherwise.
For  convenience, we  generalize bit  inversion to  an $n$-dimensional
vector $\mbf{a}$  and use an  expression $\matr{1}  - \mbf{a}$ to  denote the
$n$-dimensional vector such  that $(\matr{1} - \mbf{a})(i) =  1 - \mbf{a}(i)$
for $i\,(1\leq i \leq n)$.  $\matr{1} - \matr{A}$ is treated similarly.

\section{Computing supported models in vector spaces}\label{support}
In this  section, we  formulate the semantics  of supported  models in
vector spaces and show how to compute it by cost minimization.

\subsection{Matricized programs}\label{matprog}

\begin{definition}[Matricized program]
A  program $P$  that  has  $m$  rules  in $n$  atoms  is
numerically encoded  as a pair $\matr{P}  = (\matr{C},\matr{D})$ of binary  matrices $\matr{C} \in \{0,1\}^{m \times 2n}$ and $\matr{D} \in
\{0,1\}^{n \times m}$, which we call
a matricized program $P$.
\end{definition}

$\matr{C}$ represents rule bodies in  $P$.  Suppose atoms
are ordered like ${\cal A} = \{ a_1,\ldots,a_n \}$ and similarly rules
are ordered  like $\{ r_1: a_{i_1}  \leftarrow G_1,\ldots, r_m:a_{i_m}
\leftarrow G_m \}$.  Then the $j$-th  row $\matr{C}(j,:)$ ($1 \leq j \leq m$)
encodes  the $j$-th  conjunction  $G_j$ of  the  $j$-th rule  $a_{i_j}
\leftarrow  G_j$.  Write  $ G_j  = a_{i_1}  \wedge\cdots\wedge a_{i_p}
\wedge \neg a_{i_{p+1}} \wedge\cdots\wedge  \neg a_{i_{p+q}}$ ($1 \leq
p,q \leq n$). Then an element of $\matr{C}(j,:)$ is zero except for $\matr{C}(j,i_1)
= \cdots = \matr{C}(j,i_p) = \matr{C}(j,n+i_{p+1}) = \cdots = \matr{C}(j,n+i_{p+q}) = 1$.
$\matr{D}$ combines these  conjunctions as a disjucntion (DNF)  for each atom
in ${\cal A}$.  If the $i$-th atom  $a_i \in {\cal A}$ ($1 \leq i \leq
n$)  has  rules  $\{  a_i \leftarrow  G_{j_1},\ldots,  a_i  \leftarrow
G_{j_s} \}$  in $P$,  we put  $\matr{D}(i,j_1) =  \cdots =  \matr{D}(i,j_s) =  1$ to
represent a  disjunction $G_{j_1} \vee  \cdots \vee G_{j_s}$  which is
the right hand side of the completed rule for $a_i$: $\riff(a_i) = a_i
\Leftrightarrow G_{j_1}  \vee \cdots \vee  G_{j_s}$.  If $a_i$  has no
rule, we put $\matr{D}(i,j)  = 0$ for all $j$ ($1 \leq j  \leq m$).  Thus the
matricized  $\matr{P}  = (\matr{C},\matr{D})$  can  represent   the  completed  program
$\mbox{comp}(P)$.

For concreteness, we explain by an example below.
\begin{example}[Encoding a program]
Suppose we are given  a program
$P_0$ below containing three rules $\{r_1,r_2,r_3\}$ in a set of atoms
${\cal A}_0 = \{p,q,r\}$.

\begin{eqnarray}
P_0 & = &
     \begin{array}{ll}
     \left\{\;
          \begin{array}{lllll}
            p & \leftarrow\, & q\; \wedge\, \neg r & \hspace{1em} & \mbox{: rule $\,r_1\,$ for $p$} \\
            p & \leftarrow\, & \neg q              &              & \mbox{: rule $\,r_2\,$ for $p$} \\
            q &              &                     &              & \mbox{: rule $\,r_3\,$ for $q$} \\
          \end{array}
     \right.
     \end{array}  \label{progP0}
\end{eqnarray}
\noindent
Assuming atoms are  ordered as $p,q,r$ and correspondingly  so are the
rules $\{r_1,r_2,r_3\}$ as in (\ref{progP0}), we encode $P_0$ as a pair
of  matrices $\matr{P}_0 = (\matr{C}_0,  \matr{D}_0)$.  Here $\matr{C}_0$  represents conjunctions  (the
bodies of $\{r_1,r_2,r_3\}$) and $\matr{D}_0$ their disjunctions so that they
jointly  represent   $P_0$.

\begin{eqnarray}
\matr{C}_0\; & = &
  \begin{array}{ll}
  \begin{matrix}
    \;\;\;\; p & \;q & \;r & \!\!\!\!\neg p & \!\!\!\neg q & \!\!\!\!\neg r \\
    \vspace{-0.8em}
  \end{matrix} \\
  \begin{bmatrix}
    \;\;0 & \,1 & \,0    & \;0 & \,0  & \,1\;\; \\
    \;\;0 & \,0 & \,0    & \;0 & \,1  & \,0\;\; \\
    \;\;0 & \,0 & \,0    & \;0 & \,0  & \,0\;\; \\
  \end{bmatrix}
  &
  \begin{matrix}
    \;\; \mbox{: $r_1$ has the body $q \wedge \neg r$}\hfill \\
    \;\; \mbox{: $r_2$ has the body $\neg q$}\hfill \\
    \;\; \mbox{: $r_3$ has the empty body}\hfill \\
  \end{matrix}
  \end{array}    \label{matQ0}
\end{eqnarray}

\begin{eqnarray}
\matr{D}_0\; & = &
  \begin{array}{ll}
  \begin{matrix}
    \;\;\;\; r_1 & r_2 & \!r_3 \\
    \vspace{-0.8em}
  \end{matrix} \\
  \begin{bmatrix}
  \;\;1 & \,1 & \,0 \;\;  \\
  \;\;0 & \,0 & \,1 \;\;  \\
  \;\;0 & \,0 & \,0 \;\;  \\
  \end{bmatrix}
     &
  \begin{matrix}
    \;\; \mbox{: $p\,$ has two rules $r_1$ and $r_2$}\hfill \\
    \;\; \mbox{: $q\;$ has one rule $r_3$}\hfill \\
    \;\; \mbox{: $r\;$ has no rule}\hfill \\
  \end{matrix}
  \end{array}    \label{matD0}
\end{eqnarray}\\

As can be seen, $\matr{C}_0$ represents conjunctions in $P_0$ in such a way that
$\matr{C}_0(1,:)$ for example represents the conjunction $q \wedge \neg r$ of the first rule
in $P_0$ by setting $\matr{C}_0(1,2) = \matr{C}_0(1,6) = 1$  and so on.
$\matr{D}_0$ represents disjunctions of rule bodies. So
$\matr{D}_0(1,1) = \matr{D}_0(1,2) = 1$ means the first atom $p$ in $\{p,q,r\}$
has two rules, the first rule $r_1$ and the second rule $r_2$,
representing a disjunction $(q \wedge \neg r) \vee \neg q$ for $p$.
\end{example}

\subsection{Evaluation of formulas and the reduct of a program in vector spaces}\label{evalvec}
Here we  explain how the  propositional formulas  and the reduct  of a
program are evaluated by a model in vector spaces.  Let $I$ be a model
over a  set ${\cal A}$  of atoms.  Recall that  $I$ is equated  with a
subset of ${\cal A}$.  We  inductively define the relation ``a formula
$F$ is true in $I$'', $I \models F$ in notation, as follows.  For an atom $a$, $I \models
a$ iff $a \in  I$. For a compound formula $F$, $I  \models \neg F$ iff
$I \not\models F$.  When $F$ is a disjunction $F_1 \vee\cdots\vee F_u$
($u \geq 0$), $I \models F$ iff there  is some $i$ ($1 \leq i \leq u$)
s.t.  $I \models  F_i$.  So the empty disjunction ($u  = 0$) is always
false.  We  consider a conjunction  $F_1 \wedge\cdots\wedge F_u$  as a
syntax sugar  for $\neg (\neg  F_1 \vee\cdots\vee \neg F_u)$  using De
Morgan's law.  Consequently the empty conjunction is always true.
Let $P$ be a program having $m$  ordered rules in $n$ ordered atoms as
before  and  $G  =  a_{i_1}  \wedge\cdots\wedge  a_{i_p}  \wedge  \neg
a_{i_{p+1}} \wedge\cdots\wedge \neg a_{i_{p+q}}$ the body of a rule $a
\leftarrow G$  in $P$.  By definition,  $I \models G$ ($G$  is true in
$I$)    iff     $\{a_{i_1},\ldots,a_{i_p}\}    \subseteq     I$    and
$\{a_{i_{p+1}},\ldots,a_{i_{p+q}} \}  \cap I  = \emptyset$.   Also let
$\riff(a_i) =  a_i \Leftrightarrow G_{j_1} \vee\cdots\vee  G_{j_s}$ be
the  completed rule  for an  atom  $a_i$ in  $P$.  We  see $I  \models
\riff(a_i)$ \;iff\; $\left(  a_i \in I \;\mbox{iff}\;  I \models G_{j_1}
\vee\cdots\vee G_{j_s} \right)$.

Now we isomorphically  embed the above symbolic evaluation  to the one
in vector spaces.  Let  $I$ be a model over ordered  atoms ${\cal A} =
\{  a_1,\ldots,a_n\}$.  We  first  vectorize $I$  as  a binary  column
vector $\mbf{s}_I$  such that $\mbf{s}_I(i)  = 1$  if $a_i \in  I$ and
$\mbf{s}_I(i) =  0$ ($1 \leq i  \leq n$) otherwise, and  introduce the
dualized    $\mbf{s}_I$    written    as    $\mbf{s}_I^{\delta}$    by
$\mbf{s}_I^{\delta}        =       [\mbf{s}_I;        (\matr{1}-\mbf{s}_I)]$.
$\mbf{s}_I^{\delta}$ is  a vertical  concatenation of  $\mbf{s}_I$ and
the bit inversion of $\mbf{s}_I$.

Consider a matricized program $\matr{P} = (\matr{C},\matr{D})$ 
$( \matr{C}  \in \{0,1\}^{m \times 2n}$, $ \matr{D} \in \{0,1\}^{n \times m} )$ and its $j$-th
rule  $r_j$ having  a  body $G_j$  represented  by $\matr{C}(j,:)$.   Compute
$\matr{C}(j,:)\mbf{s}_I^{\delta}$ which is the number of true literals in $I$
in   $G_j$   and   compare   it    with   the   number   of   literals
$|\matr{C}(j,:)|_1$\footnote{
$|\mbf{v}|_1  = \sum_{i}  |\mbf{v}(i)| $  is  the 1-norm  of a  vector
  $\mbf{v}$.
} in  $G_j$.  When $|\matr{C}(j,:)|_1 =  \matr{C}(j,:)\mbf{s}_I^{\delta}$ holds, all
literals in $G_j$ are true in $I$  and hence the body $G_j$ is true in
$I$.  In  this way, we  can algebraically  compute the truth  value of
each  rule body,  but since  we consider  a conjunction  as a  negated
disjunction, we  instead compute  $\matr{C}(j,:)(\matr{1}-\mbf{s}_I^{\delta})$ which
is the number of false literals  in $G_j$. If this number is non-zero,
$G_j$ have at least one literal false in $I$,  and hence $G_j$ is  false in $I$.
The converse is also true.  The  existence of a false literal in $G_j$
is thus  computed by $\rmin_1(\matr{C}(j,:)(\matr{1}-\mbf{s}_I^{\delta}))$  which is
$1$ if there is a false literal, and $0$ otherwise.  Consequently $1 -
\rmin_1(\matr{C}(j,:)(\matr{1}-\mbf{s}_I^{\delta}))  =  1$  if  there  is  no  false
literal   in  $G_j$   and  vice   versa.    In  other   words,  $1   -
\rmin_1(\matr{C}(j,:)(\matr{1}-\mbf{s}_I^{\delta}))$ computes $I \models G_j$.

Now let  $\{ a_i\leftarrow G_{j_1},\ldots, a_i  \leftarrow G_{j_s} \}$
be   an  enumeration   of   rules   for  $a_i   \in   {\cal  A}$   and
$G_{j_1}\vee\cdots\vee G_{j_s}$ the disjunction of the rule bodies.  $
d_i =  \sum_{t=1}^s (1- \rmin_1(\matr{C}(j_t,:)(\matr{1}-\mbf{s}_I^{\delta})))  $ is
the  number of  rule bodies  in $\{G_{j_1},\ldots,G_{j_s}\}$  that are
true in  $I$.  Noting $\matr{D}(i,j) =  1$ if $j \in  \{j_1,\ldots,j_s\}$ and
$\matr{D}(i,j)  = 0$  otherwise by  construction of  $\matr{D}$ in  $\matr{P} =  (\matr{C},\matr{D})$, we
replace  the summation  $\sum_{t=1}^s$ by  matrix multiplication  and
obtain $ d_i = \matr{D}(i,:)(\matr{1}- \rmin_1(\matr{C}(\matr{1}-\mbf{s}_I^{\delta}))) $.
Introduce a column vector $\mbf{d}_I =  \matr{D}(\matr{1}- \rmin_1(\matr{C}(\matr{1}-\mbf{s}_I^{\delta})))$.
We have  $\mbf{d}_I(i) = d_i$ = the number of rules for $a_i$
whose bodies are true in $I$ ($1 \leq i \leq n$).

In the case of $\matr{P}_0 =  (\matr{C}_0,\matr{D}_0)$ in (\ref{progP0}) having three rules
$\{r_1,r_2,r_3\}$, take a model $I_0  = \{p,q\}$ over the ordered atom
set ${\cal A}_0 =  \{p,q,r\}$ where $p$ and $q$ are  true in $I_0$ but
$r$ is false in $I_0$.  Then we have
$\mbf{s}_{I_0} = [\,1 \;\;1 \;\;0 \;]^T$,
$\mbf{s}_{I_0}^{\delta} = [\,1 \;\;1 \;\;0 \;\;0 \;\;0 \;\;1 \,]^T$,
$\matr{1} -\mbf{s}_{I_0}^{\delta} =  [\,0 \;\;0  \;\;1 \;\;1  \;\;1  \;\;0  \,]^T$
and  finally
$\matr{C}_0(\matr{1}-\mbf{s}_{I_0}^{\delta}) =  [\,0 \;\;1 \;\;0 \,]^T$.
The last equation says that the  rule bodies of $r_1$, $r_2$ and $r_3$
have respectively zero, one and zero literal false in $I_0$.
Hence
$\rmin_1(\matr{C}_0(\matr{1}-\mbf{s}_{I_0}^{\delta}))  =   [\,0  \;\;1  \;\;0 \,]^T$
indicates  that only the  second rule  body is false  and the
other  two   bodies  are  true   in  $I_0$.   So  its   bit  inversion
$\matr{1}-\rmin_1(\matr{C}_0(\matr{1}-\mbf{s}_{I_0}^{\delta}))  = [\,1  \;\;0 \;\;1  \,]^T$
indicates that the second rule body is false in $I_0$ while others are
true  in $I_0$.   Thus by  combining these  truth values  in terms  of
disjunctions      $D_0$,      we     obtain
$\mbf{d}_{I_0} = \matr{D}_0(\matr{1}-\rmin_1(\matr{C}_0(\matr{1}-\mbf{s}_{I_0}^{\delta}))) = [\,1 \;\;1 \;\;0 \,]^T$.
The elements in 
$\mbf{d}_{I_0} = [\,1 \;\;1 \;\;0 \,]^T$  denote for each atom $a \in
{\cal  A}_0$  the number  of  rules  for $a$  whose  body  is true  in
$I_0$. For  example $\mbf{d}_{I_0}(1) =  1$ means that the  first atom
$p$ in  ${\cal A}_0$  has one  rule ($p \leftarrow  q \wedge  \neg r$)
whose  body  ($q   \wedge  \neg  r$)  is  true   in  $I_0$.   Likewise
$\mbf{d}_{I_0}(2) = 1$ means that the second atom $q$ has one rule ($q
\leftarrow $) whose body (empty)  is true in $I_0$.  $\mbf{d}_{I_0}(3)
= 0$  indicates that the third  atom $r$ has no  such rule.  Therefore
$\rmin_1(\mbf{d}_{I_0}) =  [\,1 \;\;1  \;\;0 \,]^T$ denotes  the truth
values of  the right hand sides  of the completed rules  $\{ \riff(p),
\riff(q),\riff(r)  \}$ evaluated  by  $I_0$.\\

\begin{proposition}
\label{prop1}
Let $\matr{P}  = (\matr{C},\matr{D})$ be  a matricized  program $P$ in  a set of atoms
${\cal A}$ and $\mbf{s}_I$ a vectorized model $I$ over ${\cal A}$. Put
$\mbf{d}_I = \matr{D}(\matr{1}- \rmin_1(\matr{C}(\matr{1}-\mbf{s}_I^{\delta})))$.  It holds that
\begin{eqnarray}
I \models \mbox{comp}(P)  & \;\;\mbox{iff}\;\; &   \| \mbf{s}_I - {\rm min}_1(\mbf{d}_I) \|_2 = 0.
\end{eqnarray}
\end{proposition}
\noindent
(Proof) Put $n = |{\cal  A}|$.  Suppose $I \models \mbox{comp}(P)$ and
write $\riff(a_i)$, the completed rule for  an atom $a_i \in {\cal A}$
($1  \leq   i  \leq   n$),  as   $\riff(a_i)  =   a_i  \Leftrightarrow
G_{j_1}\vee\cdots\vee  G_{j_s}$  ($s \geq  0$).   We  have $I  \models
\riff(a_i)$.  So if $\mbf{s}_I(i) = 1$,  $I \models a_i$, and hence $I
\models G_{j_1}\vee\cdots\vee G_{j_s}$, giving $d_i \geq 1$ because $d_i$
is the number of rule  bodies in $\{G_{j_1},\ldots,G_{j_s}\}$ that are
true in $I$.  So $\rmin_1(d_i) = 1$ holds.  Otherwise if $\mbf{s}_I(i)
=   0$,   we   have   $I   \not\models   a_i$   and   $I   \not\models
G_{j_1}\vee\cdots\vee G_{j_s}$.  Consequently none of the  rule bodies
are true in $I$ and we have $d_i = \rmin_1(d_i) = 0$.  Putting the two
together, we have  $\mbf{s}_I(i) = d_i$.  Since  $i$ is arbitrary,
we  conclude $\mbf{s}_I  =  {\rm min}_1(\mbf{d}_I)$,  or equivalently  $\|
\mbf{s}_I  -  {\rm  min}_1(\mbf{d}_I)  \|_2 =  0$.   The  converse  is
similarly proved. \hspace{1em} Q.E.D.\\

Proposition~\ref{prop1} says  that whether  $I$ is  a supported
model of the program $P$ or  not is determined by computing $\mbf{s}_I
- {\rmin}_1(\mbf{d}_I)$ in  vector spaces whose complexity  is $O(mn)$
where $m$ is the  number of rules in $P$, $n$  that of atoms occurring
in $P$. 
In the case of \(\matr{P}_0 = (\matr{C}_0, \matr{D}_0)\) with \(\mbf{s}_{I_0} = [\,1 \;\;1 \;\;0 \;]^T\) as in the aforementioned example, since $\mbf{s}_{I_0} = \rmin_1(\mbf{d}_{I_0})$ holds, it follows from Proposition~\ref{prop1} that $I_0$ is a supported model of $P_0$.

We next show how  $P^{I}$, the reduct of $P$ by $I$,  is dealt with in
vector spaces.  We assume $P$ has  $m$ rules $\{ r_1,\ldots,r_m \}$ with
a set ${\cal A} = \{a_1,\ldots,a_n \}$ of $n$ ordered atoms as before.
We first show  the evaluation of the reduct of  the matricized program
$\matr{P} = (\matr{C},\matr{D})$ by a vectorized model $\mbf{s}_{I}$.
Write $\matr{C} \in \{0,1\}^{m \times 2n}$ as $\matr{C} = [\matr{C}^{pos}\; \matr{C}^{neg}]$ where
$\matr{C}^{pos}  \in \{0,1\}^{m  \times n}$  (resp.  $\matr{C}^{neg}  \in \{0,1\}^{m
  \times  n}$)  is the  left  half  (resp.   the  right half)  of  $\matr{C}$
representing the  positive literals (resp. negative  literals) of each
rule body in $\matr{C}$.   Compute $\matr{M}^{neg} = \matr{1}-\rmin_1(\matr{C}^{neg}\mbf{s}_{I})$.
It is an  $m \times 1$ matrix  (treated as a column  vector here) such
that $\matr{M}^{neg}(j) = 0$ if the body of $r_j$ contains a negative literal
false in $I$ and  $\matr{M}^{neg}(j) = 1$ otherwise ($1 \leq  j \leq m$). Let
$r_j^{+}$ be a rule $r_j$ with  negative literals in the body deleted.
We see that $P^{I} = \{ r_j^{+} \mid \matr{M}^{neg}(j)=1, 1 \leq j \leq m \}$
and $P^{I}$ is syntactically represented by $(\matr{C}^{pos},\matr{D}^I)$ where $\matr{D}^I
=  \matr{D}$ with  columns $\matr{D}(:,j)$  replaced by  the zero  column vector  if
$\matr{M}^{neg}(j) = 0$  ($1 \leq j \leq m$).  $\matr{D}^I(i,:)$  denotes a rule set
$\{ r_j^{+} \mid \matr{D}^I(i,j) = 1, 1 \leq j \leq m \}$ in $P^{I}$ for $a_i
\in {\cal A}$.  We call  $P^{I} = (\matr{C}^{pos},\matr{D}^I)$ the matricized reduct
of $P$ by $I$.

The matricized reduct  $P^{I} = (\matr{C}^{pos},\matr{D}^I)$ is  evaluated in vector
spaces   as   follows.   Compute   $\matr{M}^{pos}   =   \matr{M}^{neg}\odot  (\matr{1}   -
\rmin_1(\matr{C}^{pos}(\matr{1}-\mbf{s}_{I})))$.
% \footnote{
% %
% $\odot$ is component-wise product.
% %
% }.
$\matr{M}^{pos}$ denotes  the truth  values  of rule  bodies in  $P^{I}$
evaluated  by $I$.   Thus $\matr{M}^{pos}(j)  =  1$ ($1  \leq j  \leq m$)  if
$r_j^{+}$  is   contained  in  $P^{I}$   and  its  body  is   true  in
$I$.  Otherwise $\matr{M}^{pos}(j)  = 0$  and  $r_j^{+}$ is  not contained  in
$P^{I}$  or  the  body  of  $r_j^{+}$  is  false  in  $I$.   Introduce
$\mbf{d}_{I}^{+} = \matr{D}\matr{M}^{pos}$. $\mbf{d}_{I}^{+}(i)$ ($1 \leq i \leq n$)
is the number of rules in $P^{I}$  for the $i$-th atom $a_i$ in ${\cal
  A}$ whose bodies are true in $I$.

\begin{proposition}
\label{prop2}
Let $\matr{P}  = (\matr{C},\matr{D})$  be a  matricized program $P$  in a  set ${\cal  A} =
\{a_1,\ldots,a_n  \}$ of  $n$ ordered  atoms   and $I$  a model  over
${\cal A}$.  Write $\matr{C} = [\matr{C}^{pos}\; \matr{C}^{neg}]$ as above. Let $\mbf{s}_I$
be   the   vectorized   model   $I$.   Compute   $\matr{M}^{neg}   =   \matr{1}-{\rm
  min}_1(\matr{C}^{neg}\mbf{s}_{I})$,  $\matr{M}^{pos}  =  \matr{M}^{neg}\odot  (\matr{1}  -  {\rm
  min}_1(\matr{C}^{pos}(\matr{1}-\mbf{s}_{I})))$  and $\mbf{d}_{I}^{+}  = DM^{pos}$.
Also compute $\mbf{d}_I = \matr{D}(\matr{1}- {\rm min}_1(\matr{C}(\matr{1}-\mbf{s}_I^{\delta})))$.
Then,
$I \models {\rm comp}(P)$,
$\| \mbf{s}_I - {\rm min}_1(\mbf{d}_I) \|_2 = 0$,
$\| \mbf{s}_I - {\rm min}_1(\mbf{d}_I^{+}) \|_2 = 0$ and
$I \models {\rm comp}(P^{I})$
are all equivalent. (Proof in Appendix \ref{sec:appdx_prop2}.)
\end{proposition}
%
% (Proof sketch) 
% (Proof) We  prove $\mbf{d}_I  = \mbf{d}_I^{+}$  first.  Recall  that a
% rule $r_j^{+}$  in $P^{I}$  is created  by removing  negative literals
% true in $I$ from the body of $r_j$  in $P$.  So for any $a_i \in {\cal
%   A}$, it is immediate that $a_i$ has a rule $r_j \in P$ whose body is
% true in $I$ iff  $a_i$ has the rule $r_j^{+} \in  P^{I}$ whose body is
% true in $I$.  Thus $\mbf{d}_I(i) = \mbf{d}_I^{+}(i)$ for every $i$ ($1
% \leq i \leq n$), and hence $\mbf{d}_I = \mbf{d}_I^{+}$.  Consequently,
% we  have $\|  \mbf{s}_I -  {\rm min}_1(\mbf{d}_I)  \|_2 =  0$ iff  $\|
% \mbf{s}_I -  {\rm min}_1(\mbf{d}_I^{+})  \|_2 =  0$.  Also  $I \models
% \mbox{comp}(P^I)$ iff $\| \mbf{s}_I  - {\rm min}_1(\mbf{d}_I^{+}) \|_2
% =  0$ is  proved similarly  to {\bf  Proposition}~\ref{prop1} (details
% omitted).         The       rest        follows       from        {\bf
%   Proposition}~\ref{prop1}. \hspace{1em} Q.E.D.\\
% %     Na = Q2*a;                       % if row in Q2 is zero vector -> Na=0 -> Ma=1
% %     Ma = 1 - (Na>=1);                % negative bodies evaluated by a
% %     GLa = Ma.*(1 - min(Q1*(1-a),1)); % Ma = indicator of survived bodies in nQ
% %     DGLa = (D*GLa)>=1;               % if a has no disjunct in D -> a is false
% %     error_fp = sum(abs(a - DGLa));   % l1_norm error

From the  viewpoint of  end-to-end ASP,  Proposition~\ref{prop2}
means that  we can obtain a  supported model $I$ as  a binary solution
$\mbf{s}_I$ of the  equation $\mbf{s}_I - {\rm  min}_1(\mbf{d}_I) = 0$
derived  from $P$  or $  \mbf{s}_I -  {\rm min}_1(\mbf{d}_I^{+})  = 0$
derived from the reduct $P^I$.   Either equation is possible and gives
the  same result  but their  computation will  be different.   This is
because the  former equation  $\mbf{s}_I - {\rm  min}_1(\mbf{d}_I)$ is
piecewise linear  w.r.t.~ $\mbf{s}_I$ whereas the  latter $\mbf{s}_I -
{\rm   min}_1(\mbf{d}_I^{+})$    is   piecewise    quadratic   w.r.t.~
$\mbf{s}_I$.\\

%
%I is a stable model of P => I is minimal in the set of {J | J|=comp(P)}
%
%Suppose I is a stable model of P.
%Consider I,J |= comp(P) such that J=<I.
%I,J |= comp(P)  =>  I|=comp(P^I)>= lfp(P^I), J|=comp(P^J)>= lfp(P^J) and
%          J=<I  =>  P^I =< P^J  =>  lfp(P^I) =< lfp(P^J)
%So J >= lfp(P^J) >= lfp(P^I) = I
%Hence, J = I, i.e.,I is minimal.
%
%I |= comp(P) and not stable
%<=> I |=comp(P^I) and not stable
%<=> I > lfp(P^I)  and I|=comp(P)
%=> I|=a s.t. a not in lfp(P^I)
%             => from I|=comp(P) and I|=a, there is a rule a:-G in P,I|=G
%             => there is a rule a:-G^+ in P^I
%=>  a:-b... in P^I and I|=b and b not in lfp(P^I)
%=> I >= loop L={a,b..} in P^I
%
% Although there are many characterizations of stable model semantics \cite{Lifschitz08b},
% we inroduce here yet another charcterization.

\begin{example}[Evaluation of a reduct]
Now look  at $P_0 =  \{r_1,r_2,r_3 \}$  in (\ref{progP0}) and  a model
$I_0 = \{p,q\}$ again.
$
\begin{array}{lcl}
P_0^{I_0} & = &
     \begin{array}{ll}
     \left\{\;
          \begin{array}{lllll}
            p & \leftarrow\, & q \\
            q & \leftarrow\, &   \\
          \end{array}
     \right.
     \end{array}  \label{progP0I0}
\end{array}
$
is the  reduct of  $P_0$ by  $I_0$.  $P_0^{I_0}$  has the  least model
$\{p,q\}$ that  coincides with $I_0$.  So  $I_0$ is a stable  model of
$P_0$.
To simulate the reduction process of  $P_0$ in vector spaces, let $\matr{P}_0
= (\matr{C}_0,\matr{D}_0)$  be the  matricized $P_0$.  We  first decompose  $\matr{C}_0$ in
(\ref{matQ0}) as $\matr{C}_0 = [\matr{C}_0^{pos} \; \matr{C}_0^{neg}]$ where $\matr{C}_0^{pos}$ is
the positive  part and $\matr{C}_0^{neg}$  the negative part of  $\matr{C}_0$.  They
are

\begin{eqnarray*}
\hspace{2em}
\begin{array}{lcl}
\matr{C}_0^{pos} & = &
  \begin{bmatrix}
    \;\;0 & \,1 & \,0 \;\; \\
    \;\;0 & \,0 & \,0 \;\; \\
    \;\;0 & \,0 & \,0 \;\; \\
  \end{bmatrix}
\end{array}
\hspace{1em}\mbox{and}\hspace{1em}
\begin{array}{lcl}
\matr{C}_0^{neg} & = &
  \begin{bmatrix}
    \;\;0 & \,0 & \,1 \;\; \\
    \;\;0 & \,1 & \,0 \;\; \\
    \;\;0 & \,0 & \,0 \;\; \\
  \end{bmatrix}
\end{array}.
\end{eqnarray*}

Let $\mbf{s}_{I_0} = [\,1 \;\;1  \;\;0 \,]^T$ be the vectorized $I_0$.
We  first compute  $\matr{M}_0^{neg} =  \matr{1}-\rmin_1(\matr{C}_0^{neg}\mbf{s}_{I_0})$ to
determine rules  to be removed.   Since $\matr{M}_0^{neg} = [\,1  \;\;0 \;\;1  \,]^T$,
the  second rule $r_2$,  indicated by $\matr{M}_0^{neg}(2) =  0$, is
removed from $P_0$, giving $P_0^{I_0} = \{ r_1^{+},r_3^{+} \}$.
Using $\matr{M}_0^{neg}$  and $\matr{D}_0$ shown  in (\ref{matD0}), we  then compute
$\matr{M}_0^{pos} =  \matr{M}_0^{neg}\odot (\matr{1}-\rmin_1(\matr{C}_0^{pos}(\matr{1}-\mbf{s}_{I_0}))) =
[\,1 \;\;0 \;\;1 \,]^T$ and  $ \mbf{d}_{I_0}^{+} = \matr{D}_0\matr{M}_0^{pos} = [\,1
  \;\;1 \;\;0  \,]^T $.   The elements of $\mbf{d}_{I_0}^{+}$ denote the  number of rule
bodies  in $P_0^{I_0}$  that are true  in  $I_0$ for  each  atom.  Thus,  since
$\mbf{s}_{I_0}  =  \rmin_1(\mbf{d}_{I_0}^{+})   (=  [\,1  \;\;1  \;\;0
  \,]^T)$  holds,  $I_0$ is  a  supported  model  of $P_{0}$  by 
  Proposition~\ref{prop2}.
\end{example}

\subsection{Cost minimization for supported models}\label{supported}

Having  linear algebraically  reformulated several  concepts in  logic
programming,  we  tackle the  problem  computing  supported models  in
vector  spaces.   Although there  already  exist  approaches for  this
problem, we tackle it without assuming any condition on programs while
allowing constraints.  Aspis et  al.~formulated the problem as solving
a non-linear  equation containing a sigmoid  function \cite{Aspis2020}.
They  encode normal  logic  programs differently  from  ours based  on
Sakama's  encoding  \cite{Sakama17} and  impose  the  MD condition  on
programs  which is  rather restrictive.   No support  is provided  for
constraints in their approach.  Later Takemura and Inoue proposed another
approach \cite{Takemura2022}  which encodes  a program  in terms  of a
single  matrix  and  evaluates  conjunctions by  the  number  of  true
literals.  They compute supported  models by minimizing a non-negative
function,  not  solving  an  equation  like  \cite{Aspis2020}.   Their
programs are however  restricted to those satisfying  the SD condition
and constraints are not considered.

Here we introduce  an end-to-end way of computing  supported models in
vector spaces through  cost minimization of a new  cost function based
on the evaluation  of disjunction. We impose  no syntactic restriction
on programs and allow constraints.  We believe that these two features
make our end-to-end ASP approach more feasible.\\

We  can   base  our  supported   model  computation  either   on 
 Proposition~\ref{prop1}  or on  Proposition~\ref{prop2}.   In
the  latter case,  we  have  to compute  GL  reduction which  requires
complicated computation compared to the  former case.  So for the sake
of simplicity, we explain the former.
Then our task in vector spaces  is to find a binary vector $\mbf{s}_I$
representing a supported model $I$ of a matricized program $\matr{P} =(\matr{C},\matr{D})$
that satisfies $ \| \mbf{s}_I - {\rm min}_1(\mbf{d}_I) \|_2 = 0$ where
$\mbf{d}_I = \matr{D}(\matr{1}-  \rmin_1(\matr{C}(\matr{1}-\mbf{s}_I^{\delta})))$.  For this task,
we relax $\mbf{s}_{I} \in \{0,1\}^n$ to $\mbf{s} \in \mathbb{R}^n$ and
introduce a non-negative cost function $L_{SU}$:
\begin{eqnarray}
L_{SU} & = & 0.5 \cdot \left( \| \mbf{s} - {\rm   min}_1(\mbf{d}) \|_2^2 + \ell_2 \cdot \| \mbf{s} \odot (\matr{1}- \mbf{s}) \|_2^2 \right)
    \;\;\;\;\mbox{where}\;\;\ \ell_2 > 0  \;\mbox{and}\; \mbf{d} = \matr{D}(\matr{1}- \rmin_1(\matr{C}(\matr{1}-\mbf{s}^{\delta}))).
    \label{eq:su}
\end{eqnarray}

\begin{proposition}
\label{prop3}
Let $L_{SU}$ be defined from a program $\matr{P} = (\matr{C},\matr{D})$ as above.

$L_{SU} = 0$ \hspace{0.1em} iff \hspace{0.3em} $\mbf{s}$ is
a binary vector representing a supported model of $P$.
\end{proposition}
\noindent
(Proof)  Apparently if  $L_{SU}  =  0$, we  have  $\|  \mbf{s} -  {\rm
  min}_1(\mbf{d})  \|_2^2 =  0$  and $\|  \mbf{s}  \odot (\matr{1}-  \mbf{s})
\|_2^2 = 0$.  The second equation means $\mbf{s}$ is binary ($x(1-x) =
0 \Leftrightarrow x  \in \{0,1\}$), and the first  equation means this
binary $\mbf{s}$ is a vector representing  a supported model of $P$ by
 Proposition~\ref{prop1}.  The converse is obvious.  \hspace{1em}
Q.E.D.\\

$L_{SU}$ is  piecewise differentiable  and we  can obtain  a supported
model of $P$ as a root $\mbf{s}$ of $L_{SU}$ by minimizing $L_{SU}$ to
zero using  Newton's method.   The Jacobian $J_{a_{SU}}$  required for
Newton's method  is derived as follows.   We assume $P$ is  written in
$n$    ordered    atoms    $\{a_1,\ldots,a_n\}$   and    $\mbf{s}    =
[u_1,\ldots,u_n]^{T}$ represents  their continuous truth  values where
$\mbf{s}(p) =  s_p \in \mathbb{R}$  is the continuous truth  value for
atom $a_p$ ($1 \leq p \leq n$).  For the convenience of derivation, we
introduce    the     dot    product     $(\matr{A}    \bullet     \matr{B})    =
\sum_{i,j}\matr{A}(i,j)\matr{B}(i,j)$  of matrices  $\matr{A}$  and $\matr{B}$  and a  one-hot
vector  $\mbf{I}_p$ which  is  a  zero vector  except  for the  $p$-th
element and $\mbf{I}_p(p) = 1$.   We note $(\matr{A} \bullet (\matr{B} \odot \matr{C}))
=  ((\matr{B}\odot \matr{A})   \bullet  \matr{C})$  and  $(\matr{A}   \bullet  (\matr{B}\matr{C}))  =
((\matr{B}^T \matr{A}) \bullet \matr{C}) = ((\matr{A} \matr{C}^T) \bullet \matr{B})$ hold (see Appendix \ref{sec:appdx_matrix_notation} for details).

Let $\matr{P} = (\matr{C},\matr{D})$ be the matricized program and write $\matr{C} = [\matr{C}^{pos}\,\matr{C}^{neg}]$.
Introduce $\matr{N}$, $\matr{M}$, $\mbf{d}$, $\matr{E}$, $\matr{F}$  and compute $L_{SU}$  by

\begin{eqnarray}
\begin{array}{lclll}
      \matr{N} & =\;\; & \matr{C}(\matr{1}-\mbf{s}^{\delta}) = \matr{C}^{pos}(\matr{1}-\mbf{s})+\matr{C}^{neg}\mbf{s}
                                              & & \mbox{: (continuous) counts of false literals in the rule bodies} \\
      \matr{M} & =\;\; & \matr{1} - {\rm min}_1(\matr{N})              & & \mbox{: (continuous) truth values of the rule bodies}  \\
\mbf{d} & =\;\; & \matr{D}\matr{M}                              & & \mbox{: (continuous) counts of true disjuncts for each atom}  \\
      \matr{E} & =\;\; & {\rm min}_1(\mbf{d}) - \mbf{s}  & & \mbox{: error between the estimated truth values of atoms and $\mbf{s}$}  \\
      \matr{F} & =\;\; & \mbf{s} \odot (\matr{1} - \mbf{s}) & & \mbox{: (continuous) 0 iif $\matr{s}$ is binary } \\
    L_{sq} & =\;\; & (\matr{E} \bullet \matr{E}) \\
    L_{nrm} & =\;\; & (\matr{F} \bullet \matr{F}) \\
    L_{SU} & =\;\; & 0.5\cdot(L_{sq} + \ell_2 \cdot L_{nrm}).
\end{array}  \label{eq:jsu}
\end{eqnarray}\\

%  f_sq = 0.5*( E <> E )
%  f_nrm = 0.5*( u.*(1-u) <> u.*(1-u) )
%

We then compute  the Jacobian $J_{a_{SU}}$  of $L_{SU}$ as follows (full derivation in Appendix \ref{sec:appdx_supported}):
\begin{eqnarray}
J_{a_{SU}}
 & = & \left( \frac{\partial L_{sq}}{\partial \mbf{s}} \right)
          + \ell_2 \cdot  \left( \frac{\partial L_{nrm}}{\partial \mbf{s}} \right)  \nonumber\\
 & = & (\matr{C}^{pos}-\matr{C}^{neg})^T ([\matr{N} \leq 1] \odot (\matr{D}^T ([\mbf{d} \leq 1] \odot \matr{E}))) - \matr{E}  + \ell_2 \cdot ((\matr{1} - 2\mbf{s}) \odot \matr{F})      \label{eq:jasu}\\
 &   & \;\mbox{where}\;\;
          \matr{N} = \matr{C}(\matr{1}-\mbf{s}^{\delta}),\,
          \mbf{d} = \matr{D}(\matr{1} - {\rm min}_1(\matr{N})),\,
          \matr{E} = {\rm min}_1(\mbf{d}) - \mbf{s},\,
          \;\mbox{and}\; \matr{F} = \mbf{s} \odot (\matr{1} - \mbf{s}).   \nonumber
\end{eqnarray}
Note that since \(\mbf{s}\) is a vector, the Jacobian in this case is also a vector.

\subsection{Adding constraints}\label{constraints}
A rule which has no head like $\leftarrow a \wedge \neg b$ is called a
constraint.   We  oftentimes  need   supported  models  which  satisfy
constraints.   Since constraints  are just  rules without  a head,  we
encode constraints as  rule bodies in a program using  a binary matrix
$\widehat{\matr{C}}  =  [\widehat{\matr{C}}^{pos} \, \widehat{\matr{C}}^{neg}]$.    We  call  $\widehat{\matr{C}}$  {\em  constraint
  matrix\/}.   We introduce  $\matr{N}_{\widehat{\matr{c}}}$, a  non-negative function  $L_{\widehat{\matr{c}}}$ of
$\mbf{s}$ and $L_{\widehat{\matr{c}}}$'s Jacobian $J_{a_{\widehat{\matr{c}}}}$ as follows (derivation in Appendix \ref{sec:appdx_constraints}): 

% J_constraint: \widehat{\matr{C}} = [Q1_c Q2_c] where \widehat{\matr{C}} represents constraints such as :- a&b&~c
\begin{eqnarray}
\matr{N}_{\widehat{\matr{c}}}  & = &  \widehat{\matr{C}}(\matr{1}-\mbf{s}^{\delta}) = \widehat{\matr{C}}^{pos}(\matr{1}- \mbf{s}) + \widehat{\matr{C}}^{neg}\mbf{s} \hspace{7.7em}\mbox{: number of literals falsified by \(\matr{s}\)} \nonumber \\
L_{\widehat{\matr{c}}}  & = & (\mbf{1} \bullet (\matr{1} - {\rm min}_1(\matr{N}_{\widehat{\matr{c}}}))) \hspace{0.5em}\mbox{where $\mbf{1}$ is an all-ones vector} \hspace{0.5em}\mbox{: counts of violated constraints} \label{eq:jc} \\
J_{a_{\widehat{\matr{c}}}} & = & (\widehat{\matr{C}}^{pos} - \widehat{\matr{C}}^{neg})^T [\matr{N}_{\widehat{\matr{c}}} \leq 1] \hspace{11.3em}\mbox{: the Jacobian of \(L_{\widehat{\matr{c}}}\)} \label{eq:jac}
\end{eqnarray}

The meaning of $\matr{N}_{\widehat{\matr{c}}}$ and $L_{\widehat{\matr{c}}}$  is clear when $\mbf{s}$ is binary.
Note that  any binary $\mbf{s}$  is considered as  a model over  a set
${\cal A}  = \{a_1,\ldots,a_n\}$  of $n$ ordered  atoms in  an obvious
way.  Suppose $k$  constraints are given to be  satisfied.  Then $\widehat{\matr{C}}$
is a $k \times  2n$ binary matrix and $\matr{N}_{\widehat{\matr{c}}}$ is a  $k \times 1$ matrix.
$\matr{N}_{\widehat{\matr{c}}}(i)$ ($1  \leq i \leq k$)  is the number of  literals falsified by
$\mbf{s}$ in a conjunction $G_i$  of the $i$-th constraint $\leftarrow
G_i$.  So  $\matr{N}_{\widehat{\matr{c}}}(i) = 0$,  or equivalently $1-{\rm min}_1(\matr{N}_{\widehat{\matr{c}}}(i))  = 1$
implies $G_i$  has no false  literal i.e., $\mbf{s} \models  G_i$, and
vice  versa.  Hence  $L_{\widehat{\matr{c}}} =  \sum_{i=1}^k (1-{\rm  min}_1(\matr{N}_{\widehat{\matr{c}}}(i))) =
(\mbf{1}  \bullet  (\matr{1}  -  {\rm min}_1(\matr{N}_{\widehat{\matr{c}}})))$  equals  the  number  of
violated constraints.  Consequently when  $\mbf{s}$ is binary,  we can
say that $L_{\widehat{\matr{c}}} = 0$ iff all constraints are satisfied by $\mbf{s}$.

When  $\mbf{s}$   is  not   binary  but  just   a  real   vector  $ \mbf{s} \in
\mathbb{R}^n$,  $\matr{N}_{\widehat{\matr{c}}}$ and  $L_{\widehat{\matr{c}}}$ are  thought to  be a  continuous
approximation  to  their  binary  counterparts.  Since  $L_{\widehat{\matr{c}}}$  is  a
piecewise  differentiable  non-negative  function  of  $\mbf{s}$,  the
approximation error can  be minimized to zero by  Newton's method using
$J_{a_{\widehat{\matr{c}}}}$   in  (\ref{eq:jac}).

\subsection{An algorithm for computing supported models with constraints}\label{algo}
Here  we present  a  minimization algorithm  for computing  supported
models of the matricized program $\matr{P} = (\matr{C},\matr{D})$ which satisfy constraints
represented by a constraint matrix $\widehat{\matr{C}}$.
We first combine $L_{SU}$ and $L_{\widehat{\matr{c}}}$ into $L_{SU+\widehat{\matr{c}}}$
using $\ell_3 > 0$.
\begin{eqnarray}
L_{SU+\widehat{\matr{c}}}
   & = &  L_{SU} + \ell_3\cdot L_{\widehat{\matr{c}}}      \label{eq:jsuc} \\
   & = &  0.5\cdot \left( \| \mbf{s} - {\rm  min}_1(\mbf{d}) \|_2^2 + \ell_2\cdot \| \mbf{s} \odot (\matr{1} - \mbf{s}) \|_2^2 \right)
             + \ell_3\cdot (\mbf{1} \bullet (\matr{1} - {\rm min}_1( \widehat{\matr{C}}(1-\mbf{s}^{\delta}) )))   \hspace{1em} \ell_2>0,\; \ell_3>0    \nonumber \\
   &   &  \;\mbox{where}\;\;  \mbf{d} = \matr{D}(\matr{1}- \rmin_1(\matr{C}(\matr{1}-\mbf{s}^{\delta})))                                                   \nonumber \\
J_{a_{SU+\widehat{\matr{c}}}}
  & = &  J_{a_{SU}} + \ell_3\cdot J_{a_{\widehat{\matr{c}}}}    \label{eq:jasuc}
\end{eqnarray}\\

The next proposition is immediate from Proposition ~\ref{prop3}.

\begin{proposition}
\label{prop4}
$L_{SU+\widehat{\matr{c}}} = 0$  \;iff\;\, $\mbf{s}$ represents
a supported model of $P$ satisfying a constraint matrix $\widehat{\matr{C}}$.
\end{proposition}

We compute  $L_{SU}$ in  $L_{SU+\widehat{\matr{c}}}$ by  (\ref{eq:jsu}) and  $L_{\widehat{\matr{c}}}$ by
(\ref{eq:jc}), and  their Jacobians  $J_{a_{SU}}$ and  $J_{a_{\widehat{\matr{c}}}}$ by
(\ref{eq:jasu}) and by (\ref{eq:jac}), respectively.
We minimize  the non-negative  $L_{SU+\widehat{\matr{c}}}$ to  zero by  Newton's method
using  Algorithm  1.   It finds  a  solution  $\mbf{s}_{*}$  of
$L_{SU+\widehat{\matr{c}}} =  0$ which represents  a supported model of  $P$ satisfying
constraint matrix  $\widehat{\matr{C}}$.  The  updating formula  is derived  from the
first   order  Taylor   expansion   of  $L_{SU+\widehat{\matr{c}}}$   and  by   solving
${\displaystyle L_{SU+\widehat{\matr{c}}} + (J_{a_{SU*c}}  \bullet (\mbf{s}_{\rm new} -
  \mbf{s})) =  0 }$  w.r.t.  $\mbf{s}_{\rm  new}$.  
The updating formula with a learning rate $\alpha>0$ is thus defined as follows:

\begin{eqnarray}
\mbf{s}_{\rm new} & = & \mbf{s} - \alpha \left( \frac{L_{SU+\widehat{\matr{c}}}}{ (J_{a_{SU+\widehat{\matr{c}}}} \;\bullet\; J_{a_{SU+\widehat{\matr{c}}}}) } \right) J_{a_{SU+\widehat{\matr{c}}}}
   \label{eq:update}
\end{eqnarray}

\begin{algorithm}[htbp]
\caption{ minimizing $L_{SU+\widehat{\matr{c}}}$ to zero }
\label{alg:suc}
%\setstretch{1.2}
\DontPrintSemicolon
\nl \KwIn{ matricized program $\matr{P} = (\matr{C},\matr{D})$, constraint matrix $\widehat{\matr{C}}$, $max\_itr \in \mathbb{Z}$,  $max\_try \in \mathbb{Z}$ }
\nl \KwOut{ binary vector $\mbf{s}_{*}$ representing a supported model of $P$ satisfying constraints represented by $\widehat{\matr{C}}$ }
\nl  $\mbf{s} \gets$  random initialization \\
\nl \For{$i \gets 1$ \textbf{to} max\_try}{
\nl \For{$j \gets 1$ \textbf{to} max\_itr}{
\nl    optimally threshold $\mbf{s}$ to a binary vector $\mbf{s}_{*}$ so that \\
\nl    $error \gets \| \mbf{s}_{*} - {\rm  min}_1(\mbf{d}_{*}) \|_2^2 + (\mbf{1} \bullet (\matr{1}-{\rm min}_1( \widehat{\matr{C}}(\matr{1}-\mbf{s}_{*}^{\delta}) )))$ \\
\nl    is minimum where $\mbf{d}_{*} = \matr{D}(\matr{1}- \rmin_1(\matr{C}(\matr{1}-\mbf{s}_{*}^{\delta}))$ \\
\nl  \If{ error = 0}{ {\bf break} }
\nl  Update $\mbf{s}$ by (\ref{eq:update})
     }
\nl  \If{ error = 0}{ {\bf break} }
\nl  perturbate $\mbf{s}$ to escape from a local minimum
     }
\nl \Return{$\mbf{s}_{*}$}
\end{algorithm}

Algorithm 1 is a double  loop algorithm where the inner $j$-loop
updates $\mbf{s}  \in \mathbb{R}^n$ repeatedly to  minimize $L_{SU+\widehat{\matr{c}}}$
while  thresholding   $\mbf{s}$  into  a  binary   solution  candidate
$\mbf{s}_{*} \in \{0,1\}^{n}$ for $L_{SU+\widehat{\matr{c}}}  = 0$.  The outer $i$-loop
is for retry when the inner loop fails to find a solution.
The initialization  at line 3  is carried out by  sampling $\mbf{s}(i)
\sim {\cal N}(0,1) + 0.5$ ($1 \leq i \leq n$) where ${\cal N}(0,1)$ is
the standard normal distribution.
Lines 6,7 and 8 collectively  perform thresholding of $\mbf{s}$ into a
binary $\mbf{s}_{*}$.   As the inner loop  repeats, $L_{SU+\widehat{\matr{c}}}$ becomes
smaller and  smaller and  so do $L_{sq}$  and $L_{nrm}$  in $L_{SU}$.
$L_{sq}$ being small means $\mbf{s}$ is close to a supported model of
$P$ while  $L_{nrm}$ being  small means each  element of  $\mbf{s}$ is
close  to $\{0,1\}$.   So  binarization $\mbf{s}_{*}  = [\mbf{s}  \geq
  \theta]$ with an appropriate threshold $\theta$\footnote{
Currently   given   $\mbf{s}$,   we   divide   the   interval   $[{\rm
    min}(\mbf{s}),{\rm  max}(\mbf{s})]$ into  20 equally  distributed
notches and use each notch as a threshold value $\theta$.
} has a good chance of  yielding a binary $\mbf{s}_{*}$ representing a
supported model of $P$ satisfying constraints represented by $\widehat{\matr{C}}$.
It may happen that  the inner loop fails to find  a solution.  In such
a case, we retry another $j$-loop with perturbated $\mbf{s}$ at line 12.
There $\mbf{s}$ is perturbated by  $\mbf{s} \gets 0.5(\mbf{s} + \Delta
+ 0.5)$ where $\Delta \sim {\cal N}(0,1)$ before the next $j$-loop.

\section{Computing stable models in vector spaces}\label{stable}

\subsection{Loop formulas and stable models}\label{loopf}
Let $\matr{P} = (\matr{C},\matr{D})$ be a matricized program  in a set of atoms ${\cal A} =
\{a_1,\ldots,a_n\}$   having   $m$   rules  $\{   a_{i_1}   \leftarrow
G_1,\ldots,a_{i_m} \leftarrow  G_m \}$ where $\matr{C} \in \{0,1\}^{m \times  2n}$ 
and  $\matr{D}  \in \{0,1\}^{n \times m}$.  We assume atoms  and rules
are ordered as indicated.

Computing  a supported  model of  $P$ is  equivalent to  computing any
binary fixedpoint  $\mbf{s} \in  \{0,1\}^{n} $  such that  $\mbf{s} =
{\rm  min}_1(\matr{D}(\matr{1}- \rmin_1(\matr{C}(\matr{1}-\mbf{s}^{\delta}))))$  in vector  spaces
and  in  this  sense,  it is  conceptually  simple  (though  NP-hard).
Contrastingly since stable  models are a proper  subclass of supported
models,  if  one wishes  to  obtain  precisely stable  models  through
fixedpoint  computation,   the  exclusion  of  non-stable   models  is
necessary.   Lin-Zhao's  theorem \cite{Lin04}  states  that  $I$ is  a
stable model of $P$ iff $I$ is a supported model of $P$ and
satisfies a  set of formulas  called {\em loop  formula\/}s associated
with $P$.

Let  $S  =  \{h_1,\ldots,h_p\}  \subseteq  {\cal  A}$  be  a  loop  in
$P$. Recall that $S$ is a set of atoms which are strongly connected in
the positive dependency graph of $P$\footnote{
In  the case  of a  singleton  loop $S  =  \{ h  \}$, we  specifically
require, following \cite{Lin04}, that $h$  has a self-loop, i.e.,there
must be a rule of the form $h \leftarrow h \wedge H$ in $P$.
}. A support rule for $h$ with respect to \(S\) is  a rule $h \leftarrow H$ such that $H^{+}
\cap S = \emptyset$.  $H$ is called a support body for $S$.  Introduce
a 
(conjunctive) loop formula for $S$ by

\begin{eqnarray}
LF(S) & = & (h_1  \wedge\cdots\wedge  h_p) \rightarrow  (H_1 \vee\cdots\vee H_q) \;\;\;
    \mbox{where $\{ H_1,\ldots,H_q \}$  are support bodies for $S$}.  \label{eq:LFL}
\end{eqnarray}

\noindent
Then define  {\em loop formulas  associated with $P$\/} as
$LF(P) =  \{ LF(S) \mid  \mbox{$S$ is a  loop in $P$}  \}$, 
which is treated as the conjunction of its elements.
We note that in the original form \cite{Lin04}, the antecedent of $LF(S)$ is a disjunction $(h_1\vee\cdots\vee h_p)$.
Later it was shown that the disjunctive and conjunctive loop formulas are equivalent \cite{Ferraris06},
and we choose to use the conjunctive form of $LF(S)$ as it is easier to satisfy using our method.

We evaluate $LF(P)$ by a real vector $\mbf{s} \in \mathbb{R}^{n}$.
Introduce an external support matrix $\matr{E}_{sup} \in \{0,1\}^{n \times m}$ by
$\matr{E}_{sup}(i,j) = 1$  if there is a support rule  $a_i \leftarrow G_{j}$ for
$a_i \in {\cal  A}$, else $\matr{E}_{sup}(i,j) = 0$ ($1\leq  i\leq n, 1\leq j\leq
m$).
Suppose there are  $w$ loops $\{S_1,\ldots,S_w\}$ in  $P$.  Introduce a
loop  matrix $\matr{L}_{oop}  \in  \{0,1\}^{w  \times m}$  such  that
$\matr{L}_{oop}(v,j) =  1$ if the $v$-th  loop $S_v$ has $G_j$  as a support
body for $S_v$,  else $\matr{L}_{oop}(v,j) = 0$ ($1 \leq  v \leq w$). 

\begin{example}[Encoding loop formulas]
Suppose we are given a program \(P_{L0}\):
\begin{eqnarray}
P_{L0} & = &
     \begin{array}{ll}
     \left\{\;
          \begin{array}{lllll}
            p & \leftarrow\, & q\; \wedge\, \neg r & \hspace{1em} & \mbox{: rule $\,r_1\,$ for $p$} \\
            p & \leftarrow\, & \neg s         &              & \mbox{: rule $\,r_2\,$ for $p$} \\
            q & \leftarrow\, & p              &              & \mbox{: rule $\,r_3\,$ for $q$} \\
            r & \leftarrow\, & r              &              & \mbox{: rule $\,r_4\,$ for $r$} \\
          \end{array}
     \right.
     \end{array}  \label{progPL0}
\end{eqnarray}

This program contains two loops: \(S_1 = \{p, q\}\) and \(S_2 = \{r\}\).
In this case, only \(S_1\) has an external support body \(\neg s\).
Thus, the external support matrix \(\matr{E}_{sup0}\) and the loop matrix \(\matr{L}_{oop0}\) for this program are as follows:
\begin{eqnarray}
\matr{E}_{sup0}\; & = &
  \begin{array}{ll}
  \begin{matrix}
    \;\;\;\; r_1 & \!r_2 & \!r_3 & \! r_4 \\
    \vspace{-0.8em}
  \end{matrix} \\
  \begin{bmatrix}
    \;\;0 & \,1 & \,0 & \;0  \\
    \;\;0 & \,0 & \,0 & \;0  \\
    \;\;0 & \,0 & \,0 & \;0  \\
    \;\;0 & \,0 & \,0 & \;0  \\
  \end{bmatrix}
  &
  \begin{matrix}
    \;\; \mbox{: $p\,$ has the body of \(r_2\) as the support body}\hfill \\
    \;\; \mbox{: $q\,$ has no external support}\hfill \\
    \;\; \mbox{: $r\,$ has no external support}\hfill \\
    \;\; \mbox{: $s\,$ is not part of any loops}\hfill \\
  \end{matrix}
  \end{array}    \label{matEs0}
\end{eqnarray}
\begin{eqnarray}
\matr{L}_{oop0}\; & = &
  \begin{array}{ll}
  \begin{matrix}
    \;\;\;\; r_1 & r_2 & \!r_3 \\
    \vspace{-0.8em}
  \end{matrix} \\
  \begin{bmatrix}
  \;\;0 & \,1 & \,0 \;\;  \\
  \;\;0 & \,0 & \,0 \;\;  \\
  \end{bmatrix}
     &
  \begin{matrix}
    \;\; \mbox{: \(S_1\,\) has the body of \(r_2\) as its support body}\hfill \\
    \;\; \mbox{: \(S_2\,\) has no support bodies}\hfill \\
  \end{matrix}
  \end{array}    \label{matLoop0}
\end{eqnarray}\\

\end{example}

We then introduce a loss function $L_{LF}$, which is a non-negative piecewise linear function of $\mbf{s}$.

\begin{eqnarray}
\matr{M}     & =\; &  \matr{1} - {\rm min}_1(\matr{C}(\matr{1}-\mbf{s}^{\delta})) \hspace{4em} \mbox{: (continuous) truth values by $\mbf{s}$ of the rule bodies in $P$}  \nonumber \\
\matr{S}_v   & =\; &  \matr{L}_{oop}(v,:)                          \hspace{8.6em} \mbox{: represents the $v$-th loop in $\{\matr{S}_1,\ldots,\matr{S}_t\}$}        \nonumber \\
A_v   & =\; &  \matr{S}_v(\matr{1}-\mbf{s}) + \matr{S}_v \matr{E}_{sup} \matr{M}  \hspace{4em} \mbox{: (continuous) counts of true disjuncts  by $\mbf{s}$ of $LF(S_v)$ } \nonumber \\
L_{LF} & =\; &  \sum_{v=1}^{w} \left(1 - {\rm min}_1(A_v) \right)                                    \label{eq:jLF}
\end{eqnarray}
%\begin{eqnarray}
%\begin{array}{lclll}
%M      & =\; &  1 - {\rm min}_1(C(1-\mbf{s}^{\delta}))
%S_s    & =\; &  \matr{L}_{oop}(s,:)
%A_s    & =\; &  S_s(1- \mbf{s}) + S_s\matr{E}_{sup}M
%L_{LF} & =\; &  \sum_{v=1}^{t} (1 - {\rm min}_1(A_s))
%\end{array}  \label{eq:jLFu}
%\end{eqnarray}

\begin{proposition}
\label{prop5}
Let $L_{LF}$ be  defined as above. When $\mbf{s}$ is a binary vector representing a model \(I\) over \(\mathcal{A}\), it holds that
 $L_{LF} = 0$ \;iff\; $\mbf{s} \models LF(P)$.
\end{proposition}

\noindent
(Proof) Suppose $L_{LF} = 0$ and $\mbf{s}$ is binary.  A summand $(1 -
{\rm  min}_1(A_v))$  in  $L_{LF}$ (\ref{eq:jLF})  corresponds  to  the
$v$-th loop $S_v = \{h_1,\ldots,h_p \}$ and is non-negative.  Consider
$LF(S_v)   =    (h_1   \wedge\cdots\wedge   h_p)    \rightarrow   (H_1
\vee\cdots\vee H_q)$  as a  disjunction $\neg h_1  \vee\cdots\vee \neg
h_p \vee H_1 \vee\cdots\vee H_q$.
Then  $L_{LF}  =  0$  implies  $(1   -  {\rm  min}_1(A_v))  =  0$,  or
equivalently $A_v  \geq 1$. Consequently,  as $\mbf{s}$ is  binary, we
have $\matr{S}_v(\matr{1}-\mbf{s})  \geq 1$ or  $\matr{S}_v\matr{E}_{sup} \matr{M} \geq 1$.  The  former means
$ I  \models  \neg h_1  \vee\cdots\vee  \neg  h_p$.  The  latter,
$\matr{S}_v\matr{E}_{sup} \matr{M} \geq  1$, means $  {I} \models H_1 \vee\cdots\vee  H_q $.
This is  because the element $(\matr{E}_{sup} \matr{M})(i)$ is the  number of support rules  for $a_i
\in {\cal A}$ whose bodies are true  in \(I\) $\mbf{s}$ ($1 \leq i \leq n$),
and hence  $\matr{S}_v\matr{E}_{sup} \matr{M} \geq 1$ means  some support body $H_r$  ($1 \leq r
\leq q$) for  $S_v$ is true in ${I}$.  So  in either case $I
\models LF(S_v)$.   Since $v$ is  arbitrary, we have  $ I \models
LF(P)$.  The  converse is  straightforward and  omitted.  \hspace{1em}
Q.E.D.\\

%
% N = Q1*(1-u) + Q2*u   -->  dN/du_r = Q2*1_r - Q1*1_r = (Q2-Q1)*I_r
% M = 1 - min_1(N)      -->  dM/du_r = -[N=<1].*((Q2-Q1)*I_r) = [N=<1].*((Q1-Q2)*I_r)
% Ns = L(s,:)*(1-u)     -->  dNs/du_r = -L(s,:)*I_r        -- scalar
% Ms = min(Ns,1)        -->  dMs/du = -[Ns=<1]*(L(s,:)^T)  -- (n x 1)

%--(LF_OR)(LF_AND)
% As = Ms + L(s,:)*Es*M                         -- scalar, M evaluates full bodies
% L = \matr{L}_{oop}
% Es = \matr{E}_{sup}
%Ja_{LF} = dJ_LF/du = sum_s d(1-LF*(s))/du = - sum_s dLF*(s)/du
%   = - sum_s dmin(As,1)/du
%   = - sum_s [As=<1]*( dMs/du + (L(s,:)*Es*(dM/du))^T )
%   = - sum_s [As=<1]*( -[Ns=<1]*(L(s,:)^T) + { L(s,:)*Es*(-repmat([N=<1],1,n).*(Q2-Q1)) }^T )
%   = sum_s [As=<1]*( [Ns=<1]*(L(s,:)^T) + (L(s,:)*Es*(repmat([N=<1],1,n).*(Q2-Q1)))^T )
%   = sum_s [As=<1]*( [Ns=<1]*(L(s,:)^T) + ( ((L(s,:)*Es).*[N=<1]^T)*(Q2-Q1) )^T )
%

The Jacobian $J_{a_{LF}}$ of $L_{LF}$ is computed as follows (derivation in Appendix \ref{sec:appdx_loopformula}):
\begin{eqnarray}
\matr{N}   & = & \matr{C}(\matr{1}-\mbf{s}^{\delta})  \nonumber \\
N_v & = & \matr{S}_v(\matr{1}-\mbf{s}) \nonumber \\
M_v & = & {\rm min}_1(N_v)  \nonumber \\
J_{a_{LF}}
   & = & \frac{\partial L_{LF}}{\partial \mbf{s}}
     =   \sum_{v=1}^{w} - \left( \frac{\partial {\rm min}_1(A_v)}{\partial \mbf{s}} \right) \nonumber \\
   % & = & - \sum_{v=1}^{w} [A_v \leq 1]
   %       \left( \left( \frac{\partial M_v}{\partial \mbf{s}} \right)
   %             + S_v\matr{E}_{sup}\left( \frac{\partial M}{\partial \mbf{s}} \right)^T
   %       \right)                                                                  \nonumber \\
   & = & \sum_{v=1}^{w} [A_v \leq 1]
         \left(   [N_v \leq 1]\matr{S}_v^T + ( ((\matr{S}_v\matr{E}_{sup})\odot[\matr{N} \leq 1]^T)(\matr{C}^{neg}-\matr{C}^{pos})^T
         \right)  \label{eq:jaLF}
\end{eqnarray}

\noindent
Here $\matr{C} = [\matr{C}^{pos}\, \matr{C}^{neg}]$ and
$\matr{S}_v$, $A_v$ and $\matr{M}$ are computed by (\ref{eq:jLF}).\\

Now  introduce a  new cost  function $L_{SU+\widehat{\matr{c}}+LF}$  by (\ref{eq:jsucLF})
that incorporates $ L_{LF}$ and compute its Jacobian $J_{a_{SU+\widehat{\matr{c}}+LF}}$
by (\ref{eq:jasuc}).

\begin{eqnarray}
L_{SU+\widehat{\matr{c}}+LF}
  & = & L_{SU+\widehat{\matr{c}}} +\ell_{4}\cdot L_{LF}  \;\;\mbox{where\; $\ell_{4}>0$}  \label{eq:jsucLF}  \\
%  & = &  0.5\cdot \left( \| \mbf{s} - {\rm  min}_1(\mbf{d}) \|_2^2 + \ell_2\cdot \| \mbf{s} \odot (1 - \mbf{s}) \|_2^2 \right)  \nonumber \\
%  &   &  \; + \ell_3\cdot (\mbf{1} \bullet (1 - {\rm min}_1( \widehat{\matr{C}}(1-\mbf{s}^{\delta}) )))   \hspace{1em} \ell_2>0, \ell_3>0     \nonumber \\
J_{a_{SU+\widehat{\matr{c}}+LF}}
  & = & J_{a_{SU+\widehat{\matr{c}}}} +\ell_{4}\cdot J_{a_{LF}}   \label{eq:jasucLF}
\end{eqnarray}\\

By combining Proposition ~\ref{prop4}, \ref{prop5} and Lin-Zhao's
theorem \cite{Lin04}, the following is obvious.\\

\begin{proposition}
\label{prop6}

$\mbf{s}$ is a stable model of $P$ satisfying constraints represented by $\widehat{\matr{C}}$
\;iff\;\; $\mbf{s}$ is a root of $L_{SU+\widehat{\matr{c}}+LF}$.
\end{proposition}

We compute such  $\mbf{s}$ by Newton's method using  Algorithm 1
with a modified update rule (\ref{eq:update}) such that $L_{SU+\widehat{\matr{c}}}$ and
$J_{a_{SU+\widehat{\matr{c}}}}$  are replaced  by  $L_{SU+\widehat{\matr{c}}+LF}$ and  $J_{a_{SU+\widehat{\matr{c}}+LF}}$
respectively.

When a  program $P$  is tight \cite{Fages94},  for example  when rules
have no  positive literal in their  bodies, $P$ has no  loop and hence
$LF$ is empty.  In such a case, we directly minimize $L_{SU+\widehat{\matr{c}}}$ instead
of using $L_{SU+\widehat{\matr{c}}+LF}$ with the empty $LF$.

%%=================================
\subsection{LF heuristics}\label{heuristics}
Minimizing $L_{SU+\widehat{\matr{c}}+LF}$ is  a general way of  computing stable models
under constraints.   It is applicable  to any  program and gives  us a
theoretical framework for computing stable  models in an end-to-end way
without  depending   on  symbolic  systems.   However   there  can  be
exponentially many  loops and  they make  the computation  of $L_{LF}$
(\ref{eq:jLF})  extremely  difficult  or practically  impossible.   To
mitigate this seemingly insurmountable  difficulty, we propose two
heuristics which use a subset of loop formulas.

\begin{itemize}
\item[{\bf  $LF_{max}$}:   ]  The   first  heuristic   is  $LF_{max}$.   We consider  only a  set $LF_{max}$  of loop  formulas
  associated with  SCCs in the  positive dependency graph  $\rpdg(P) =
  (V,E)$ of a program  $P$. In the case of a  singleton SCC \(\{a\}\), \(a\)
  must  have   a  self-loop  in   $\rpdg(P)$.   We  compute   SCCs  in
  $O(|E|+|V|)$ time by Tarjan's algorithm \cite{Tarjan72}.\\

\item[  {\bf  $LF_{min}$}:  ]  In this  heuristic,  instead  of  SCCs
  (maximal strongly  connected subgraphs), we choose  minimal strongly
  connected subgraphs,  i.e., cycle graphs.  Denote  by $LF_{min}$ the
  set of loop formulas associated with cycle graphs in $\rpdg(P)$.  We
  use  an   enumeration  algorithm  described  in   \cite{Liu2006}  to
  enumerate cycles and construct $LF_{min}$ due to its simplicity.\\

%\item[  {\bf $LF_{min\_max}$}:  ] The  last  one is  a combination  of
%  $LF_{max}$  and  $LF_{min}$.    Let  $(a_1  \wedge\cdots\wedge  a_p)
%  \rightarrow  (G_1   \vee\cdots\vee  G_q)$  be  a   loop  formula  in
%  $LF_{min}$. $L  = \{  a_1,\ldots,a_p \}$ is  a cycle  in $\rpdg(P)$.
%  Note that $S$ has a unique  loop (SCC) $L{ a'_1,\ldots,a'_r \}$
%  containing $S$  and a  loop formula $(a'_1  \wedge\cdots\wedge a'_r)
%  \rightarrow (H_1 \vee\cdots\vee H_t)$  in $LF_{max}$ associated with
%  $L'$.  We  consider a new implication  $(a_1 \wedge\cdots\wedge a_p)
%  \rightarrow (H_1  \vee\cdots\vee H_t)$  and call it  the synthesized
%  loop formula associated  with $S$.  Let $LF_{min\_max}$ be  a set of
%  synthesized loop formulas associated with cycles in $\rpdg(P)$.
\end{itemize}

We remark  that although $LF_{max}$  and $LF_{min}$ can  exclude some
of non-stable  models,  they  do  not  necessarily  exclude  all  of
non-stable  models.   However,  the  role  of  loop  formulas  in  our
framework is entirely different from the one in symbolic ASP.  Namely,
the role of LF in our  framework is not to logically reject non-stable
models but to  guide the search process by  their gradient information
in  the  continuous  search  space.  Hence,  we  expect,  as  actually
observed in experiments  in the next section, some  loop formulas have
the power of guiding the search process to a root of $L_{SU+\widehat{\matr{c}}+LF}$.
%
%The  last  heuristics, $LF_{min\_max}$,  is  very  different from  the
%previous two.  It is a sufficient condition for stable model. That is,
%we  can  prove  that  any   supported  model  $I$  of  $P$  satisfying
%$LF_{min\_max}$  is  a stable  model  of  $P$  (but not  vice  versa).
%Unfortunately, sometimes $LF_{min\_max}$ is  too strong and may reject all
%supported models of $P$.

\subsection{Precomputation}\label{precomp}
We introduce  here precomputation.  The  idea is to remove  atoms from
the search space  which are false in every stable  model. It downsizes
the program and realizes faster model computation.

When a program $P$ in a set ${\cal A}$ = atom($P$) is given, we transform
$P$ to  a definite program  $P^{+}$ by removing all  negative literals
from the rule bodies in $P$.  Since $P^{+} \supseteq P^{I}$ holds as a set
of  rules  for  any  model   $I$,  we  have  $LM(P^{+})  \supseteq
LM(P^{I})$ where $LM(P)$ denotes the least model of a definite
program $P$.   When $I$ is a  stable model, $LM(P^{I}) =  I$ holds
and we have $LM(P^{+}) \supseteq  I$. By taking the complements of
both  sides,  we   can  say  that  if  an  atom   $a$  is  outside  of
$LM(P^{+})$, i.e.,if $a$ is false in $LM(P^{+})$, so is $a$ in
{\em any\/} stable model $I$ of  $P$.  Thus, by precomputing the least
model $LM(P^{+})$,  we can  remove a  set of  atoms ${\cal  F}_P =
{\cal A}\setminus  LM(P^{+})$ from  our consideration as  they are
known to  be false  in any  stable model.  We  call ${\cal  F}_P$ {\em
stable false atom\/}s.
Of  course,  this  precomputation   needs  additional  computation  of
$LM(P^{+})$ but it can be done  in linear time proportional to the
size of  $P^{+}$, i.e., the  total number  of occurrences of  atoms in
$P^{+}$ \cite{Dowling84}\footnote{
We implemented  the linear  time algorithm in  \cite{Dowling84} linear
algebraically using vector and matrix and confirmed its linearity.
}.  Accordingly  precomputing the least model  $LM(P^{+})$ makes sense
if the  benefit of removing stable  false atoms from the  search space
outweighs linear time computation for  $LM(P^{+})$, which is likely to
happen when we  deal with programs with positive literals  in the rule
bodies.\\

More concretely, given  a program $P$ and a set of constraints $K$, we can
obtain downsized ones, $P'$ and $K'$, as follows.

\begin{itemize}
\item[ {\bf Step 1}: ] Compute the least model $LM(P^{+})$ and the set of
    stable false atoms ${\cal F}_P = \mbox{atom}(P)\setminus LM(P^{+})$.
\item[ {\bf Step 2}: ]  Define
   \begin{eqnarray}
     G' & = &  \mbox{conjunction $G$ with negative literals $ \{ \neg a \in G  \mid a \in {\cal F}_P \}$  removed} \nonumber \\
     P' & = &  \{ a \leftarrow G' \mid a \leftarrow G \in P, a \not\in {\cal F}_P, G^{+}\cap{\cal F}_P  = \emptyset \}
               \;\;\mbox{where}\;\; G^{+} = \mbox{positive literals in $G$}                                     \label{eq:P'}  \\
     K' &  = & \{ \leftarrow G' \mid  \leftarrow G \in K, G^{+}\cap{\cal F}_P  = \emptyset \}                  \label{eq:C'}
   \end{eqnarray}
\end{itemize}

\begin{proposition}
\label{prop7}
Let  $P'$ and  $K'$ be  respectively the  program (\ref{eq:P'})  and
constraints (\ref{eq:C'}).   Also let  $I'$ be  a model  over
atom($P'$).  Expand  $I'$ to  a model $I$  over atom($P$)  by assuming
every atom in ${\cal F}_P$ is false in $I$. Then\\

$I'$ is a stable model of $P'$ satisfying constraints $K'$
\;\:iff\;\;   $I$ is a stable model of $P$ satisfying constraints $K$.
\end{proposition}
\noindent
% $LM(P'^{I'}) \subseteq LM(P^{I})$.
% a<- G'^+ in P'^{I'}  =>  a<-G' in P' and I'|=G'^-  =>  a<-G is in P and G^+ has no f_atom and G'^+ = G^+ b.c. G'=(G\~f_atoms)
% I'|= G'^-  =>  I|= G^- b.c. (G'^- = (G^-\~f_atoms) and I|= ~f-atoms by construction (I = I'+{~f-atoms}))
%      =>  a<-G^+ in P^{I}  =>  a<-G'^+ in P^{I} b.c. G'^+ = G^+
%      =>  P'^{I'} =< P^{I} as set  =>  lfp(P'^{I'}) =< lfp(P^{I})
% a in lfp(P^{I})  =>  b<-G^+ in a proof from P^{I}, b<-G in P, I|= G^-
% P^{I} =< P^+ as set  =>  lfp(P^{I}) =< lfp(P^+)  =>  lfp(P^{I}) contains no f-atom => G^+ has no f-atom  => G'^+ = G^+
% I|= G^-  =>  I'|=G'^- b.c. G'^- = G^-\~f_atoms  =>  b<-G'^+ in P'^{I'}  =>  b<-G^+ in P'^{I'} b.c. G'^+ = G^+
%       => a is in lfp(P'^{I'}) and a is arbitrary =>  lfp(P^{I}) =< lfp(P'^{I'})
% So lfp(P^{I}) = lfp(P'^{I'}) for any I and I'
(Proof) We prove  first $I'$ is a stable model  of $P'$ iff
$I$ is a stable model of  $P$.  To prove it, we prove $LM(P'^{I'}) = LM(P^{I})$  as set.

Let    $a   \leftarrow    G'^{+}$    be   an    arbitrary   rule    in
$P'^{I'}$. Correspondingly there  is a rule $a \leftarrow  G'$ in $P'$
such that $I' \models G'^{-}$.  So there is a rule $a \leftarrow G$ in
$P$ such that  $G' = G \setminus \{  \neg b \mid b \in  {\cal F}_P \}$
and $G^{+} \cap  {\cal F}_P = \emptyset$. $I'  \models G'^{-}$ implies
$I \models G^{-}$  by construction of $I$ from $I'$.  So $a \leftarrow
G^{+}$ is contained  in $P^{I}$, which means $a  \leftarrow G'^{+}$ is
contained in  $P^{I}$ because $G'^{+}  = G^{+}$  (recall that $G'  = G
\setminus \{ \neg  b \mid b \in  {\cal F}_P \}$ and $G'$  and $G$ have
the same set of positive  literals).  Thus since $a \leftarrow G'^{+}$
is an arbitrary rule, we conclude $P'^{I'} \subseteq P^{I}$, and hence
$LM(P'^{I'}) \subseteq LM(P^{I})$.

Now  consider $a  \in  LM(P^{I})$.   There is  an  SLD derivation  for
$\leftarrow a$ from $P^{I}$.  Let $b  \leftarrow G^{+} \in P^{I}$ be a
rule  used  in the  derivation  which  is  derived  from the  rule  $b
\leftarrow  G  \in P$  such  that  $I  \models G^{-}$.   Since  $P^{I}
\subseteq P^{+}$,  we have  $LM(P^{I}) \subseteq LM(P^{+})$  and hence
$LM(P^{I}) \cap {\cal F}_P = \emptyset$, i.e., $LM(P^{I})$ contains no
stable false  atom.  So $b \not\in  {\cal F}_P$ and $G^{+}  \cap {\cal
  F}_P  = \emptyset$  because every  atom in  the SLD  derivation must
belong in $LM(P^{I})$.  Accordingly $b \leftarrow G' \in P'$.
On the other  hand, $I \models G^{-}$ implies $I'  \models G'^{-}$. So
$b  \leftarrow  G'$  is  in  $P'$ and  $b  \leftarrow  G'^{+}$  is  in
$P'^{I'}$.   Therefore $b  \leftarrow G^{+}$  is in  $P'^{I'}$ because
$G'^{+}  =  G^{+}$.   Thus  every  rule used  in  the  derivation  for
$\leftarrow a$  from $P^{I}$  is also a  rule contained  in $P'^{I'}$,
which means $a  \in LM(P'^{I'})$.  Since $a$ is  arbitrary, it follows
that  $LM(P^{I})  \subseteq  LM(P'^{I'})$.   By  putting  $LM(P'^{I'})
\subseteq LM(P^{I})$  and $LM(P^{I}) \subseteq  LM(P'^{I'})$ together,
we conclude $LM(P^{I}) = LM(P'^{I'})$.

Then, if $I'$ is  a stable model of $P'$, we have  $I' = LM(P'^{I'}) =
LM(P^{I})$ as set.  Since $I = I'$  as set, we have $I = LM(P^{I})$ as
set, which means $I$ is a stable model of $P$.  Likewise when $I$ is a
stable model of  $P$, we have $I  = LM(P^{I}) = LM(P'^{I'})$  and $I =
I'$ as set.  So  $I' = LM(P'^{I'})$ as set and $I'$  is a stable model
of  $P'$.  

As for the constraints, consider a constraint \(\leftarrow G'\) in \(K'\).
We consider two cases, where \(b \in {\cal F}_P\) occurs positively and negatively in \(G\) and \(G'\).
In the former case, the body remains the same between \(G'\) and \(G\), thus if \(I' \models G'\) then \(I \models G\) and vice versa.
In the latter case, because \(\neg b\) always evaluates to true, the negative occurrence \(\neg b\) in the body of the constraint will not change the result of the conjunction \(G\).
Thus, combining the two cases and since the constraint \(\leftarrow G'\) is arbitrary, we conclude that \(I' \models K'\) iff \(I \models K\). \hspace{1em} Q.E.D.

\section{Programming  examples}\label{example}
In this section, we  apply our ASP approach to examples  as a proof of
concept   and  examine   the  effectiveness   of  precomputation   and
heuristics.   Since large  scale computing  is  out of  scope in  this
paper, the program size is mostly small\footnote{
Matricized programs in this paper are  all written in GNU Octave 6.4.0
and run on a PC with  Intel(R) Core(TM) i7-10700@2.90GHz CPU with 26GB
memory.
}.

\subsection{The 3-coloring problem}\label{3coloring}
\begin{figure}[htbp]
\begin{tabular}{ccc}
  \begin{minipage}[b]{0.3\linewidth}
    \centering
    \includegraphics[keepaspectratio, scale=0.55]{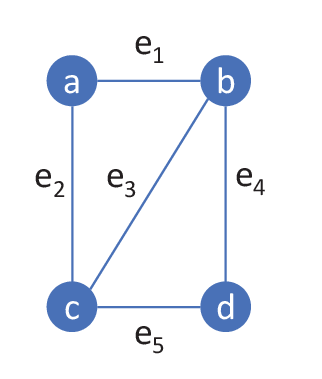}
    \subcaption{Graph $G_1$}
  \end{minipage}
& &
  \begin{minipage}[b]{0.3\linewidth}
    \centering
    \includegraphics[keepaspectratio, scale=0.55]{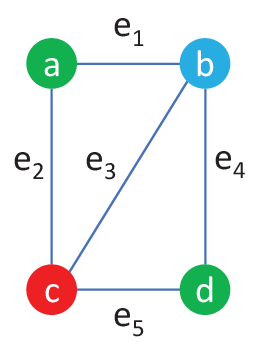}
    \subcaption{A 3-coloring}
  \end{minipage}
%
%  \begin{minipage}[b]{0.4\linewidth}
%    \centering
%    {\footnotesize
%      Program \\
%      \vspace{0.2em}
%      $ \left\{\; \begin{array}{l}
%                   a1 \leftarrow \neg a2 \wedge \neg a3,\; a2 \leftarrow \neg a3 \wedge \neg a1,  \\
%                     \hspace{1em} a3 \leftarrow \neg a1 \wedge \neg a2  \\
%                            \hspace{5em}\cdots \\
%                   d1 \leftarrow \neg d2 \wedge \neg d3,\; d2 \leftarrow \neg d3 \wedge \neg d1,  \\
%                     \hspace{1em} d3 \leftarrow \neg d1 \wedge \neg d2 \\
%                 \end{array}
%          \right.$
%    }
%    \\[1em]
%    {\footnotesize
%      Constraints \\
%      \vspace{0.2em}
%      $\left\{\; \begin{array}{l}
%                   \leftarrow a1 \wedge b1,\; \leftarrow a2 \wedge b2,\; \leftarrow a3 \wedge b3 \\
%                            \hspace{5em}\cdots \\
%                   \leftarrow d1 \wedge c1,\; \leftarrow d2 \wedge c2,\; \leftarrow d3 \wedge c3 \\
%                 \end{array}
%      \right.$
%    }
%    \vspace{1em}
%    \subcaption{Program and constraints}
%  \end{minipage}

\end{tabular}
\caption{3-coloring problem}
\label{graph:G1}
\end{figure}

We first  deal with  the 3-coloring  problem. Suppose  we are  given a
graph $G_1$.  The task is to color the vertices of the graph blue, red
and  green so  that  no two  adjacent vertices  have  the same  color
like (b) in Fig.~\ref{graph:G1}.

There are four nodes $\{a, b, c, d\}$ in the graph $G_1$.  We assign a
set of three color atoms (Boolean variables) to each node to represent
their  color.  For  example, node  $a$ is  assigned three  color atoms
$\{a_1({\rm red}),  a_2({\rm blue}), a_3({\rm green})\}$.   We need to
represent two facts by these atoms.

\begin{itemize}
\item Each node  has a unique color  chosen from $\{red,blue,green\}$.
  So color atoms assigned to each node are  in an XOR relation.
  We represent  this fact  by a tight  program $P_1$  below containing
  three rules for each node.

\begin{eqnarray}
P_{1} & = &
\left\{\;
\begin{array}{lllll}
  a_1 \leftarrow \neg a_2 \wedge \neg a_3, &  a_2 \leftarrow \neg a_3 \wedge \neg a_1, & a_3 \leftarrow \neg a_1 \wedge \neg a_2 \\
  b_1 \leftarrow \neg b_2 \wedge \neg b_3, &  b_2 \leftarrow \neg b_3 \wedge \neg b_1, & b_3 \leftarrow \neg b_1 \wedge \neg b_2 \\
  c_1 \leftarrow \neg c_2 \wedge \neg c_3, &  c_2 \leftarrow \neg c_3 \wedge \neg c_1, & c_3 \leftarrow \neg c_1 \wedge \neg c_2 \\
  d_1 \leftarrow \neg d_2 \wedge \neg c_3, &  d_2 \leftarrow \neg d_3 \wedge \neg d_1, & d_3 \leftarrow \neg d_1 \wedge \neg d_2
\end{array}
\right.     \label{prog:P1}
\end{eqnarray}

\item Two nodes connected by an  edge must have a different color.  We
  represent this fact in terms of constraints.

\begin{eqnarray}
K_1 & = &
\left\{\;
\begin{array}{lllll}
   \leftarrow a_1 \wedge b_1, &  \leftarrow a_2 \wedge b_2, &  \leftarrow a_3 \wedge b_3 \\
   \leftarrow a_1 \wedge c_1, &  \leftarrow a_2 \wedge c_2, &  \leftarrow a_3 \wedge c_3 \\
   \leftarrow b_1 \wedge c_1, &  \leftarrow b_2 \wedge c_2, &  \leftarrow b_3 \wedge c_3 \\
   \leftarrow b_1 \wedge d_1, &  \leftarrow b_2 \wedge d_2, &  \leftarrow b_3 \wedge d_3 \\
   \leftarrow d_1 \wedge c_1, &  \leftarrow d_2 \wedge c_2, &  \leftarrow d_3 \wedge c_3 \\
\end{array}
\right.      \label{C1}
\end{eqnarray}
\end{itemize}

Assuming an ordering of  atoms $\{ a_1,a_2,a_3,\ldots,d_1,d_2,d_3 \}$,
the normal logic program $P_1$  shown in (\ref{prog:P1}) is matricized
to  $\matr{P}_1 =  (\matr{C}_1,  \matr{D}_1)$ where  $\matr{D}_1$  is a  $(12  \times 12)$  binary
identity matrix (because there are 12 atoms and each atom has just one
rule)  and  $\matr{C}_1$  is  a  $(12 \times  24)$  binary  matrix  shown  in
(\ref{mat:P1}).  Constraints listed in  (\ref{C1}) are a matricized to
a ($15  \times 12$)  constraint matrix $\widehat{\matr{C}}_{K_1}$  (\ref{mat:Qc1}).  In
(\ref{mat:P1})  and  (\ref{mat:Qc1}), $a$  for  example  stands for  a
triple $(a_1\,  a_2\, a_3)$ and $\neg  a$ for $(\neg a_1\,  \neg a_2\,
\neg a_3)$.

\begin{eqnarray}
\matr{C}_1\; & = &
  \begin{array}{ll}
    \begin{matrix}
    \;\;\;\; a & \;b & \;c  & \;d & \hspace{0.1em}\neg a & \hspace{0.1em}\neg b & \hspace{0.1em}\neg c & \hspace{0.1em}\neg d \\
    \vspace{-0.8em}
  \end{matrix} \\
  \begin{bmatrix}
          &    &    &    & \hspace{3em} \matr{H}_3 &      &     &      & \\
          &    &    &    &                  &  \matr{H}_3 &     &      & \\
          &    &    &    &                  &      & \matr{H}_3 &      & \\
          &    &    &    &                  &      &     & \matr{H}_3  &
  \end{bmatrix}
  &  \;\;\mbox{where}\;\;
           \matr{H}_3  \;=\;   \begin{bmatrix}
                           0 & 1 & 1 \\
                           1 & 0 & 1 \\
                           1 & 1 & 0 \\
                       \end{bmatrix}
  \end{array}    \label{mat:P1}
\end{eqnarray}
%H3 = [0 1 1;
%      1 0 1;
%      1 1 0];

\begin{eqnarray}
\widehat{\matr{C}}_{K_1}\; & = &
  \begin{array}{ll}
    \begin{matrix}
    \;\;\;\;\; a & \hspace{0.6em} b & \hspace{0.6em} c  & \hspace{0.5em} d  \\
    \vspace{-0.8em}
  \end{matrix} \\
  \begin{bmatrix}
       \matr{E}_3  &  \matr{E}_3  &       &     \\
       \matr{E}_3  &       &  \matr{E}_3  &     \\
            &  \matr{E}_3  &  \matr{E}_3  &     \\
            &  \matr{E}_3  &       &  \matr{E}_3 \\
            &       &  \matr{E}_3  &  \matr{E}_3 \\
  \end{bmatrix}
  &  \;\;\mbox{where}\;\;
           \matr{E}_3  \;=\;   \begin{bmatrix}
                           1 & 0 & 0 \\
                           0 & 1 & 0 \\
                           0 & 0 & 1 \\
                       \end{bmatrix}
  \end{array}    \label{mat:Qc1}
\end{eqnarray}

We run  Algorithm 1 on  program $P_1$ with constraints  $K_1$ to
find a supported model (solution) of $P_1$ satisfying $K_1$\footnote{
Since $P_1$ is a tight program, every supported model of $P_1$
is a stable model and vice versa.
}.

%http://www.yamamo10.jp/yamamoto/comp/latex/make_doc/table/table.php
\begin{table*}[htb]
\caption{Time and the number of solutions} \label{table:t1} \centering
\begin{tabular}{ccc}
\toprule
 time(s)         &  \#solutions    \\
\midrule
%\addlinespace[5mm]
   6.7(0.7)      &  5.2(0.9)       \\ % program-9 in test_asp_c.m
\bottomrule
%\hline
\end{tabular}
\end{table*}

% program-9 in test_asp_c.m
To measure time to find a model, we conduct ten trials\footnote{
One trial  consists of  $max\_itr \times max\_try$  parameter updates.
} of running Algorithm 1  with $max\_try = 20$, $max\_itr = 50$,
$\ell_2  =  \ell_3  =  0.1$  and take  the  average.   The  result  is
0.104s(0.070)\footnote{The numbers in the parentheses indicate the standard deviation.} on average.
Also to check  the ability of finding different  solutions, we perform
ten trials of Algorithm 1\footnote{
without   {\em   another    solution   constraint\/}   introduced   in
Section~\ref{hamiltonian}
}and  count  the  number  of   different  solutions  in  the  returned
solutions.  \#solutions  in Table~\ref{table:t1} is the  average of ten
such measurements. 
Considering there are six solutions and the naive implementation, the number of different solutions found by the algorithm, which was 5.2 on average, seems rather high.

\begin{figure}[bhtp]
\begin{tabular}{c}
  \begin{minipage}[b]{0.2\linewidth}
     \centering                      \includegraphics[keepaspectratio,
       scale=0.35]{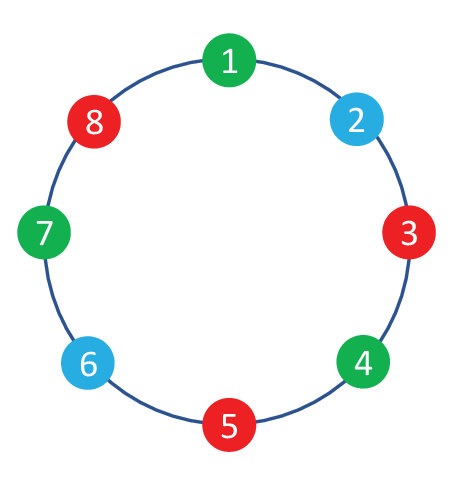}\\[2em]   \subcaption{A   cycle
       graph}
     \label{fig:cycle-1}
  \end{minipage}
  \hspace{1em}  %\hfill
  \begin{minipage}[b]{0.35\linewidth}
     \centering
     \includegraphics[keepaspectratio, scale=0.18]{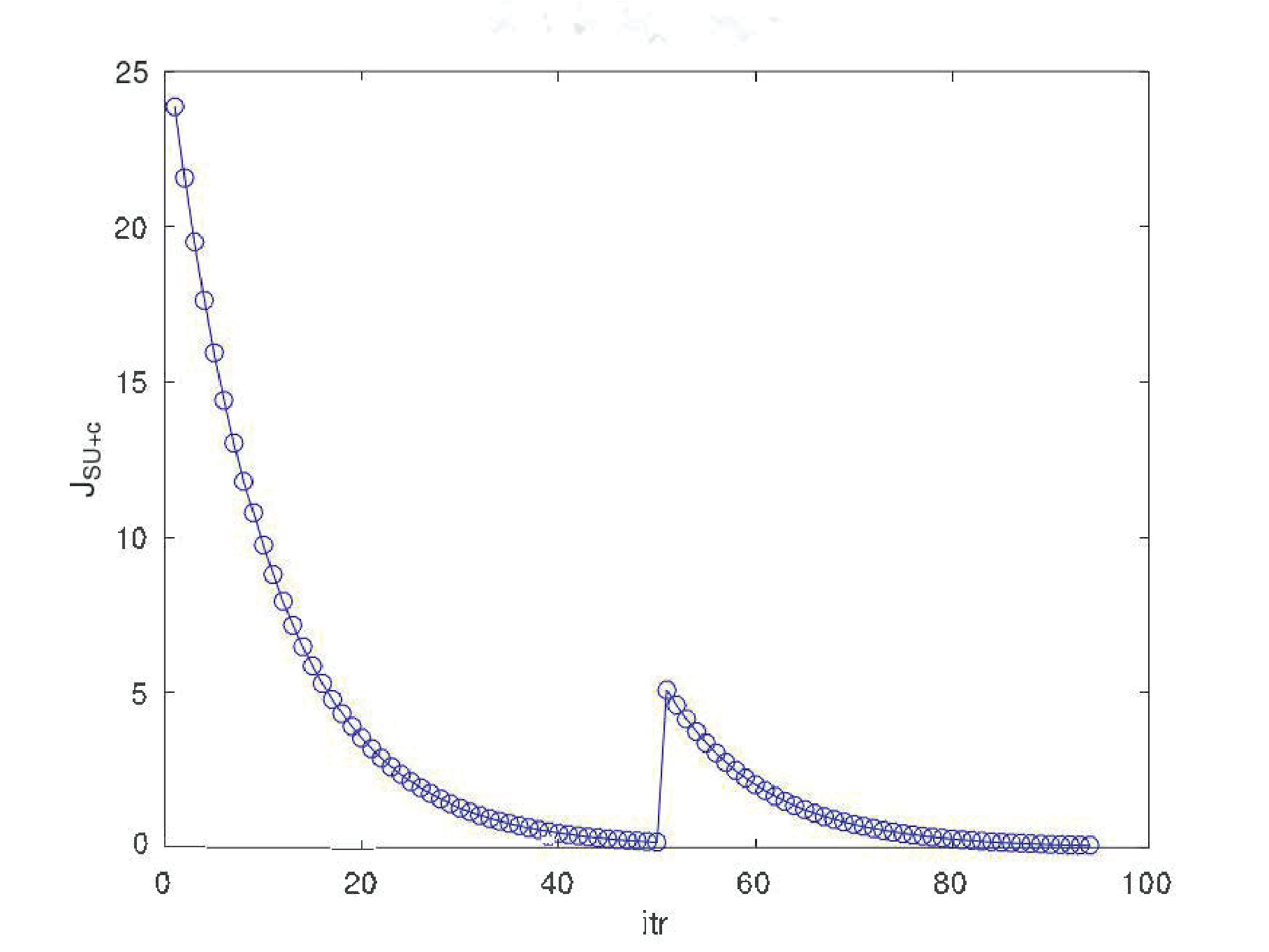}
     \subcaption{Minimization of $L_{SU+\widehat{\matr{c}}}$ with retry}
     \label{fig:cycle-2}
  \end{minipage}
%  \hspace{1em}
  \begin{minipage}[b]{0.35\linewidth}
     \centering
     \includegraphics[keepaspectratio, scale=0.45]{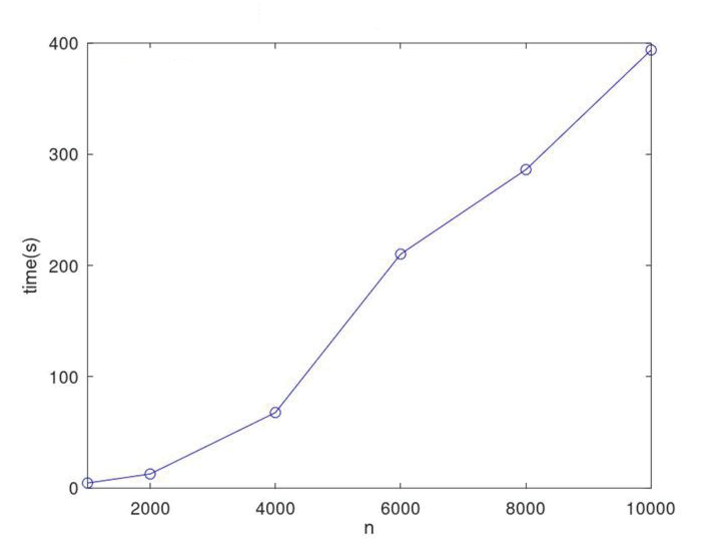}
     \subcaption{Scalability}
     \label{fig:cycle-3}
  \end{minipage}
\end{tabular}
\caption{Convergence and scalability}
\label{Ex-2:cycle-graph}
\end{figure}

% program-2 in test_asp_c.m
Next we check the scalability of our approach by a simple problem.  We
consider   the   3-coloring   of   a   cycle   graph   like   (a)   in
Fig.~\ref{Ex-2:cycle-graph}.  In general, given a cycle graph that has
$n$ nodes, we encode its 3-coloring problem as in the previous example
by  a matricized  program $\matr{P}_2  = (\matr{C}_2,\matr{D}_2)$  and a  constraint matrix
$\widehat{\matr{C}}_{K_2}$ where  $\matr{D}_2(3n \times 3n)$  is an identity matrix  and $\matr{C}(3n
\times 6n)$  and $\widehat{\matr{C}}_{K_2}(3n \times 6n)$  represent respectively rules
and constraints.  There are $2^{n} + 2(-1)^{n}$ solutions ($n \geq 3$)
in $2^{3n}$ possible  assignments for $3n$ atoms\footnote{
Based on the \textit{chromatic polynomial} for cycle graphs (for an introduction, see for example \cite{WestGraph2001}), which is given by \(P(C_n, \lambda)=(\lambda - 1)^n + (-1)^n(\lambda - 1)\) for a cycle graph \(C_n\) with \(n\) vertices (\(n \geq 3\)) colored by \(\lambda\) colors.}.
So  the problem will
be exponentially difficult as $n$ goes up.

The  graph  (b)  in   Fig.~\ref{Ex-2:cycle-graph}  is  an  example  of
convergence curve  of $L_{SU+\widehat{\matr{c}}}$ by  Algorithm  1 with $n  = 10$,
$max\_try = 100$, $max\_itr = 50$.  The  curve tells us that in the first
cycle  of $j$-loop,  the  inner  for loop  of  Algorithm 1,  no
solution  is found  after  $max\_itr  = 50$  iterations  of update  of
continuous assignment vector $\mbf{s}$.  Then perturbation is given to
$\mbf{s}$ which  causes a small jump  of $L_{SU+\widehat{\matr{c}}}$ at $itr  = 51$ and
the second cycle of $j$-loop starts  and this time a solution is found
after dozens of  updates by thresholding $\mbf{s}$ to  a binary vector
$\mbf{s}_*$.

The graph (c) in  Fig.~\ref{Ex-2:cycle-graph} shows the scalability of
computation  time  to find  a  solution  up to  $n  =  10000$. We  set
$max\_try  =  100$, $max\_itr  =  2000$  and  plot  the average  of  ten
measurements of time to find a  solution.  The graph seems to indicate
good linearity w.r.t.~$n$ up to $n = 10000$.

\subsection{The Hamiltonian cycle problem, precomputation and another solution constraint}\label{hamiltonian}
A  Hamiltonian cycle  (HC) is  a cycle  in a  graph that  visits every
vertex exactly once and the  Hamiltonian cycle problem is to determine
if an HC  exists in a given  graph.  It is an  NP-complete problem and
has been  used as a programming  example since the early  days of ASP.
Initially, it  is encoded  by a  non-tight program  containing positive
recursion \cite{Niemela99}.  Later a way of encoding by a program that
is  not  tight  but  tight   on  its  completion  models  is  proposed
\cite{Lin03}.   We here  introduce  yet another  encoding  by a  tight
ground program inspired by SAT encoding proposed in \cite{Zhou20}
where  Zhou showed  that the  problem is  solvable by  translating six
conditions listed in Fig.~\ref{SAT:HC} into a SAT problem\footnote{
Actually, \cite{Zhou20} listed seven conditions to be encoded as a SAT problem.
However, one of them is found to be redundant and we use the remaining
six conditions.
}.\\

\begin{figure}[hbtp]
\begin{tabular}{rlll}
\rule{0mm}{1.2em}
    &  \hspace{2em} conditions                   &   & \hspace{10em} meaning  \\ \hline
\rule{0mm}{1.2em}
(1) & ${\rm one\_of}(H_{i,j_1},\ldots,H_{i,j_k})$   & : & one of outgoing edges $\{ i \rightarrow j_1,\ldots,i\rightarrow j_k \}$ from
                                                       vertex $i$ is in an HC \\
(2) & $U_{j,q} \leftarrow H_{i,j} \wedge U_{i,q-1}$ & : & if edge $i \rightarrow j$ is in an HC and vertex $i$ is visited at time $q-1$, \\
    &                                            &   & vertex $j$ is visited at time $q$ $(1 \leq i,j, q \leq  N)$ \\
(3) & $U_{1,1}$                                   & : & vertex 1 is visited at time 1 (starting point) \\
(4) & ${\rm one\_of}(H_{i_1,j},\ldots,H_{i_k,j})$   & : & one of incoming edges $\{ i_1\rightarrow j,\ldots,i_k\rightarrow j \}$ to
                                                        vertex $j$ is in an HC \\
(5) & $ \leftarrow H_{i,1} \wedge \neg U_{i,N}$    & : & if edge $i \rightarrow 1$ is in an HC, vertex $i$ is visited at time $N$ $(2 \leq i \leq N)$ \\
\vspace{0.2em}
(6) & ${\rm one\_of}(U_{i,1},\ldots,U_{i,N})$      & : & vertex $i$ is visited once ($1 \leq i \leq N$) \\ \hline
\end{tabular}
\caption{Conditions for SAT encoding of a Hamiltonian cycle problem.}
\label{SAT:HC}
\end{figure}
\noindent
In what follows, we assume vertices are numbered from $1$ to $N$ = the
number of vertices in a graph.  We  use $i \rightarrow j$ to denote an
edge from  vertex $i$ to  vertex $j$  and $H_{i,j}$ to  indicate there
exists an edge from  $i$ to $j$ in an HC.   $U_{j,q}$ means vertex $j$
is  visited   at  time  $q$  ($1   \leq  j,  q  \leq   N$)  and  ${\rm
  one\_of}(a_1,\ldots,a_k)$ means  that one of $\{  a_1,\ldots,a_k \}$
is true.
We translate these  conditions into a program $P_{3}  = \{ (1),(2),(3)
\}$ and constraints $K_{3} = \{  (4),(5),(6) \}$.  To be more precise,
the first condition $(1)$ is translated  into a tight program just like
a  program $P_1$  (\ref{prog:P1}).  The  conditions $\{  (2), (3)  \}$
constitute  a  tight  definite   program.   Constraints  $K_{2}  =  \{
(4),(5),(6)  \}$ are  encoded as  a set  of implications  of the  form
$\leftarrow L_1  \wedge\cdots\wedge L_k$  where $L_1,\ldots,  L_k$ are
literals. A  set of $U_{j,q}$  atoms contained  in a stable  model of
$P_{3}$ satisfying $K_{3}$ gives an HC.

\begin{figure}[htbp]
\begin{tabular}{ccc}
  \begin{minipage}[b]{0.4\linewidth}\centering
    \includegraphics[keepaspectratio,scale=0.5]{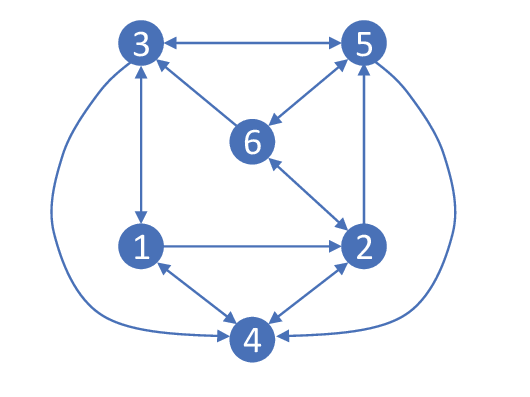}
    \subcaption{Graph $G_2$}\label{fig:3}
  \end{minipage}
& \hspace{1em} &
  \begin{minipage}[b]{0.4\linewidth}\centering
%  \begin{table*}[hb]
% {\small
  \begin{tabular}{ccc}
    \toprule
                  &  no precomp. &  precomp.  \\
    \midrule
    time(s)       &  2.08(2.01)  &  0.66(0.52)  \\
    \#solutions   &  4.9         &  5.7        \\
    \bottomrule
  \end{tabular}\\
%  }
%  \end{table*}
  \vspace{3em}
  % max_fp = 7         time(s)       |supported|=|stable|    ave.over ten runs
  % max_try = 20;  max_itr = 200;  %l2=l3=0.1 l5=0.5 for no_LF
   \subcaption{Time and the number of different solutions}
%  \subcaption{The effect of precomputation}
  \end{minipage}
\end{tabular}
\caption{A HC problem}
\label{graph:HC}
\end{figure}

We apply the above encoding to a simple Hamiltonian cycle problem for
a graph $G_2$ in Fig.~\ref{graph:HC}\footnote{
$G_2$ is taken  from: Section 6.2 in \textit{Potassco User Guide}
({\tt https://github.com/potassco/guide/releases/tag/v2.2.0}).
}. There are six vertices and six HCs\footnote{
They are
$  1 \rightarrow 2 \rightarrow 5 \rightarrow 6 \rightarrow 3 \rightarrow 4 \rightarrow 1 $,
$  1 \rightarrow 2 \rightarrow 6 \rightarrow 3 \rightarrow 5 \rightarrow 4 \rightarrow 1 $,
$  1 \rightarrow 2 \rightarrow 6 \rightarrow 5 \rightarrow 3 \rightarrow 4 \rightarrow 1 $,
$  1 \rightarrow 3 \rightarrow 5 \rightarrow 6 \rightarrow 2 \rightarrow 4 \rightarrow 1 $,
$  1 \rightarrow 4 \rightarrow 2 \rightarrow 5 \rightarrow 6 \rightarrow 3 \rightarrow 1 $,
$  1 \rightarrow 4 \rightarrow 2 \rightarrow 6 \rightarrow 5 \rightarrow 3 \rightarrow 1 $.
}. To solve this HC problem,  we matricize $P_{3}$ and $K_{3}$.  There
are 36  $H_{i,j}$ atoms ($1 \leq  i,j \leq 6$) and  36 $U_{j,q}$ atoms
($1 \leq j,q \leq  6$).  So there are 72 atoms in  total.  $P_{3} = \{
(1),(2),(3) \}$ contains 197 rules in  these 72 atoms and we translate
$P_{3}$ into  a pair of matrices  $(\matr{C}_{3}, \matr{D}_3)$ where $\matr{D}_{3}$  is a
$72 \times 197$ binary matrix for disjunctions\footnote{
For example, for  each $U_{j,q}$ ($1 \leq j,q \leq  6$), condition (2)
generates six  rules $\{  U_{j,q} \leftarrow H_{i,j}  \wedge U_{i,q-1}
\mid 1 \leq i \leq 6 \}$.
}and  $\matr{C}_3$ is  a $197  \times  144$ matrix  for conjunctions  (rule
bodies).  Likewise  $K_{3} = \{  (4),(5),(6) \}$ is translated  into a
constraint matrix $\widehat{\matr{C}}_{K_3}$ which is  a $67 \times 144$ binary matrix.
A more detailed description of the encoding is available in the appendix (Appendix \ref{sec:appdx_hamiltonian}).
Then   our  task   is  to   find  a   root  $\mbf{s}$   of  $L_{SU+\widehat{\matr{c}}}$
(\ref{eq:jsuc}) constructed from these  $\matr{C}_{3}$, $\matr{D}_3$ and $\widehat{\matr{C}}_{K_3}$
in a 72 dimensional vector space by minimizing $L_{SU+\widehat{\matr{c}}}$ to zero.

% P_HC = (D(72x197),nC(197x144)) for 72 atoms (36 H_ij's, 36 U_ij's) encodes:
%  (r1) one_of(H_{i,j1}..H_{i,jk})      :there are exclusively outgoing edges i->j1..i->jk for every node i
%  (r5) H_{i,j} & U_{i,q-1} => U_{j,q}  :if i->j exists and i is visited at time q-1 (1 =< i,j,q= < 6=|cities|)
%  (r7) U_{1,1}                         :city 1 is visited at time 1 (starting node)
% C_HC = \widehat{\matr{C}}(26x72) encodes
%  (r2) one_of(H_{i1,j}..H_{ik,j})      :there are exclusively incoming edges i1->j..ik->j for every j
%  (r4) :- H_{i,1}&~U_{i,6}             :if i->1 exists, i is visited at time 6 (2=<i=<6)
%  (r6) one_of(U_{i,1}..U_{i,6})        :city i is visited once (1=<i=<6)

% precomp_time(s) =  0.0023780
%   D:(72x197)  ->(40x90) 14184:3600
%   nQ:(197x144)->(90x80) 28368:7200
%   \widehat{\matr{C}}:(67x144)->(52x80)  9648:4160
% |asp_false_atoms|/|total atoms| = 32/72
%% (Comment):
%% <> The number of atoms in the search space is reduced from 72 to 40 by precompuation
%% <> precomputation requires 0.002s but it
%     significantly improves time and the number of solutions
%     obtained by max_fp trials of P_HC with no_LF
%     (time(s): 2.08 -> 0.66, |sol|: 4.9 -> 5.7 at max_fp = 7)

We apply  precomputation in the  previous section to $\matr{P}_{3}  = (\matr{C}_{3},
\matr{D}_3)$  and $\widehat{\matr{C}}_{K_3}$  to reduce  program size.   It takes  2.3ms and
detects  32 false  stable  atoms.  It  outputs  a precomputed  program
$\matr{P}'_{3} = (\matr{C}'_{3}, \matr{D}'_{3})$ and a constraint matrix $\widehat{\matr{C}}'_{K_3}$ of size
$\matr{D}'_{3} (40  \times 90)$,  $\matr{C}'_{3} (90 \times  80)$ and  $\widehat{\matr{C}}'_{K_3} (52
\times  80)$ respectively,  which is  $1/4$ or  $1/2$ of  the original
size.  So precomputation removes $45\%$ of atoms from the search space
and returns much smaller matrices.

We run  Algorithm 1 on  $\matr{P}_{3} = (\matr{C}_{3}, \matr{D}_3)$  with $\widehat{\matr{C}}_{K_3}$
(no  precomputation) and  also  on $\matr{P}'_{3}  =  (\matr{C}'_{3}, \matr{D}'_{3})$  with
$\widehat{\matr{C}}'_{K_3}$ (precomputation)  using $max\_try  = 20$, $max\_itr  = 200$
and  $\ell_2 =  \ell_3 =  0.1$ and  measure time  to find  a solution,
i.e., stable model  satisfying constraints.   The result  is shown  by
Table  (b)  in Fig.~\ref{graph:HC}  as  time(s)  where time(s)  is  an
average  of ten  trials.   Figures  in the  table,  2.08s vs.   0.66s\footnote{This includes the time for precomputation.},
clearly demonstrate the usefulness of precomputation.

In  addition to  computation  time,  we examine  the  search power  of
different  solutions  in  our  approach by  measuring  the  number  of
obtainable solutions.  More concretely, we run Algorithm 1 seven
times, and each time a stable  model is obtained as a conjunction $L_1
\wedge\ldots\wedge  L_{72}$  of  literals,  we add  a  new  constraint
$\leftarrow  L_1 \wedge\ldots\wedge  L_{72}$ to  previous constraints,
thereby forcibly computing a new stable  model in next trial.  We call
such use of constraint {\em another solution constraint}.  Since there
are at most  six solutions, the number of solutions  obtained by seven
trials is at  most six.  We repeat  a batch of seven  trials ten times
and  take the  average of  the number  of solutions  obtained by  each
batch.   The average  is denoted  as  \#solutions in  Table (b)  which
indicates  that 5.7  solutions out  of  6, almost  all solutions,  are
obtained by seven trials using another solution constraint.

Summing  up,  figures in  Table  (b)  exemplify the  effectiveness  of
precomputation  which  significantly   reduces  computation  time  and
returns  a  more  variety  of solutions  when  combined  with  another
solution constraint.
% max_fp = 7         time(s)       |supported|=|stable|    ave.over ten runs

\subsection{LF heuristics and precomputation on loopy programs}\label{LF_heuristics}
So far we have been dealing with tight programs which have no loop and
hence  have no  loop formulas.   We here  deal with  non-tight programs
containing  loops  and  examine  how  LF  heuristics,  $LF_{max}$  and
$LF_{min}$,
%
%  and $LF_{min\_max}$,
%
introduced  in  the  previous  section  work.  We  use  an  artificial
non-tight  program $P_{4\_n}$  (with no  constraint) shown  below that
has exponentially many loops.

%program-18   (non-tight program, no-constraint)
%-----------------
%% [ Loopy program P_loop2_4 ]:
% n = 4;  % n+2 atoms  n:even
% n:even  n+2 atoms {a0,a1,..,an,a(n+1)}
%   a0:- a1,..,an
%   a(2i-1):- a0 v a(2i)     for i=1..n/2
%   a(2i):-   a0 v a(2i-1)   for i=1..n/2
%   a(n+1):- a(n+1)
%   a0:- ~a(n+1)

% P_loop2_n has n+2 atoms,
%    2^{n/2}+1 supported models
%    1 stable model {a0,a1..an,~a(n+1)} = [1...1 0]^T

%% [ #models and LF ]
% atoms(P_loop2_n)   = n+2 atoms
% |supported models| = 2^{n/2}+1
% |stable models|    = 1 = {a0,a1..an,a(n+1)=0}
%
% LF_max: { a0&a1&..&an->~a(n+1), a(n+1)->false }                  |LF_max| = 2
%     => |supported model| = 1 allowred
% LF_min: { a1&a2->a0, a3&a4->a0,.. a(n+1)->false }                |LF_min| = n/2+1
%     => |supported model| = 1 allowred
% LF_min_max = { a1&a2->false, a3&a4->false,.. a(n+1)->false }     |LF_min_max| = n/2+1
%     => |supported model| = 0
% n = 50; max_try = 10;  max_itr = 100;  max_fp = 2**(n/2)+1;    %l3 = l2 = 0.1, l5 = 1.0

%\begin{figure}[htbp]
\begin{eqnarray*}
P_{4\_n} & = &
     \begin{array}{ll}
     \left\{\;
          \begin{array}{lllll}
            a_0      & \leftarrow\, & a_1 \wedge\ldots\wedge a_n \\
            a_{0}    & \leftarrow\, & \neg a_{n+1} \\
            \ldots  \\
            a_{2i-1}  & \leftarrow\, & a_0 \vee a_{2i}              & \mbox{for  $i:1 \leq i \leq n/2$} \\
            a_{2i}    & \leftarrow\, & a_0 \vee a_{2i-1}            & \mbox{for  $i:1 \leq i \leq n/2$} \\
            \ldots  \\
            a_{n+1}    & \leftarrow\, & a_{n+1} \\
          \end{array}
     \right.
     \end{array} %  \label{prog:loopy1}
\end{eqnarray*}
%\caption{A non-tight program $P_3$}
%\label{prog:loop1}
%\end{figure}

\noindent
We here consider an even \(n\), then
 $P_{4\_n}$   program   has    $n+2$   atoms
$\{a_0,a_1,\ldots,a_n,a_{n+1}  \}$, $2^{n/2}+1$  supported models  and
one stable model $\{a_0,a_1,\ldots,a_n\}$.   There are $n/2+1$ minimal
loops  $\{a_{1},a_{2}\},\ldots,\{a_{n-1},a_{n}\},  \{a_{n+1}\}$ and  a
maximal loop  $\{a_0,a_1,\ldots,a_n\}$.  The set of  loop formulas for
LF heuristics are computed as follows.

\begin{eqnarray*}
 LF_{max} & = & \{ (a_0 \wedge a_1 \wedge\ldots\wedge a_n) \rightarrow \neg a_{n+1},\,          a_{n+1}\rightarrow \perp \} \\
 LF_{min} & = & \{ (a_1\wedge a_2)\rightarrow a_0,\ldots,(a_{n-1}\wedge a_{n})\rightarrow a_0,\, a_{n+1}\rightarrow \perp \} \\
\end{eqnarray*}
%     b.c. a0 in the maximal loop {a0,a1..an} has an external support rule { a0:- ~a(n+1) }
%     all supported models with a(n+1)=1 excluded but those with a(n+1)=0 allowed
%     a(n+1)=0 => a0=1 => a1..an=1 => supported model {a0,a1..an} remains
%     => |stable model| = 1
%        fp_set' = asp_set' = [1 1 1 1 1 0] experimentally confirmed for n=4
%
Note that  although there are  $2^{n/2}+1$ supported models,  there is
only one stable model.  So $LF_{max}$  and $ LF_{min}$ are expected to
exclude $2^{n/2}$ supported models.

% nC(2*n+3 x 2*n+4);
% D((n+2) x (2n+3))
% \widehat{\matr{C}} = []
After  translating $P_{4\_n}$  into a  matricized program  $\matr{P}_{4\_n} =
(\matr{C}_{4\_n},  \matr{D}_{4\_n})$ where  $\matr{C}_{4\_n}$ is  a $(2n+3)  \times (2n+4)$
binary matrix and $\matr{D}_{4\_n}$ is  a $(n+2) \times (2n+3)$ binary matrix
respectively, we compute a stable  model of $P_{4\_n}$ for various $n$
by Algorithm 1  that minimizes  $L_{SU+\widehat{\matr{c}}+LF}$ (\ref{eq:jsucLF})
with coefficient $\ell_3  = 0$ for the constraint term  (because of no
use of constraints) using Jacobian $J_{a_{SU+\widehat{\matr{c}}+LF}}$ (\ref{eq:jasuc}).

Below is an example of the program \(P_{4\_n}\) where \(n = 4\).
\begin{eqnarray*}
P_{4\_4} & = &
     \begin{array}{ll}
     \left\{\;
          \begin{array}{lllll}
            a_0  & \leftarrow & a_1 \wedge a_2 \wedge a_3 \wedge a_4 \\
            a_0  & \leftarrow & \neg a_5 \\
            a_1  & \leftarrow & a_0 & a_1 \leftarrow a_2 \\
            a_2  & \leftarrow & a_0 & a_2 \leftarrow a_1 \\
            a_3  & \leftarrow & a_0 & a_3 \leftarrow a_4 \\
            a_4  & \leftarrow & a_0 & a_4 \leftarrow a_3 \\
            a_5  & \leftarrow & a_5 \\
          \end{array}
     \right.
     \end{array} %  \label{prog:loopy1}
\end{eqnarray*}
This program has three minimal loops \(\{a_1,a_2\},\{a_3,a_4\},{a_5}\) and a maximal loop \(\{a_0,a_1,a_2,a_3,a_4\}\).
There are 11 rules and six atoms, so \(\matr{C}_{4\_4}\) is a \((11 \times 12)\) binary matrix.

Since all supported  models of $P_{4\_n}$ except for  one stable model
are non-stable, even  if $LF_{max}$ and $ LF_{min}$ are  used to guide
the search process towards a stable model, Algorithm 1 is likely
to return a non-stable model.  We  can avoid such a situation by the use
of another solution constraint.

\begin{table*}[thb]
\caption{The effect of another solution constraint} \label{table:t2}
\centering
\begin{tabular}{ccc}
\toprule
another solution constraint
                   &   time(s)     &   \#trials   \\
\midrule
  not used         &  $11.46(0.41)$  &  $10^4(0)$    \\
   used            &  $0.09(0.13)$   &  $3.5(1.6)$   \\
\bottomrule
\end{tabular}
\end{table*}

To  verify  it,  we  examine  the  pure  effect  of  another  solution
constraint that guides the search process to compute a model different
from   previous  ones.    Without  using   $LF_{max}$  or   $LF_{min}$
heuristics, we  repeatedly run Algorithm  1 with/without another
solution constraint for  $10^4$ trials with $n = 4$,  $max\_try = 20$,
$max\_itr = 50$,  $\ell_2 = \ell_3 =  0.1$ and measure time  to find a
stable model  and count the  number of  trials until then.   We repeat
this experiment ten  times and take the average.  The  result is shown
in Table~\ref{table:t2}.

The figure $10^4(0)$ in Table~\ref{table:t2} in  the case of no use of
another solution  constraint means  Algorithm 1  always exhausts
$10^4$ trials without finding a stable  model (due to implicit bias in
Algorithm 1).  When another solution constraint is used however,
it finds  a stable model in  0.09s after 3.5 trials  on average.  Thus
Table~\ref{table:t2} demonstrates  the necessity and  effectiveness of
another solution constraint to efficiently explore the search space.\\

% n = 4, max_try = 20;  max_itr = 50; max_fp = 10000; %<- one run repeats 10^4 trials
%
% 1 trial = (max_itr x max_try) computation
% 1 run = max_fp trials
% time = time to find one stable model or exhaust all max_fp trials
%
% no_LF (no loop formula used) +  --(another sol.) + --(one sol.)
% ave. time over ten run with max_fp=10000
%
% (another solution)
%     constraint          time(s)      trials
% ----------------------------------------------
%     not used          11.46(0.41)    10^4(0)     <-- no stable model found after 10^4 trials
%     used               0.09(0.13)    3.5(1.6)    <-- one stable model found

\begin{figure}[htbp]
\begin{tabular}{ccc}
  \begin{minipage}[b]{0.4\linewidth}\centering
    \includegraphics[keepaspectratio,scale=0.27]{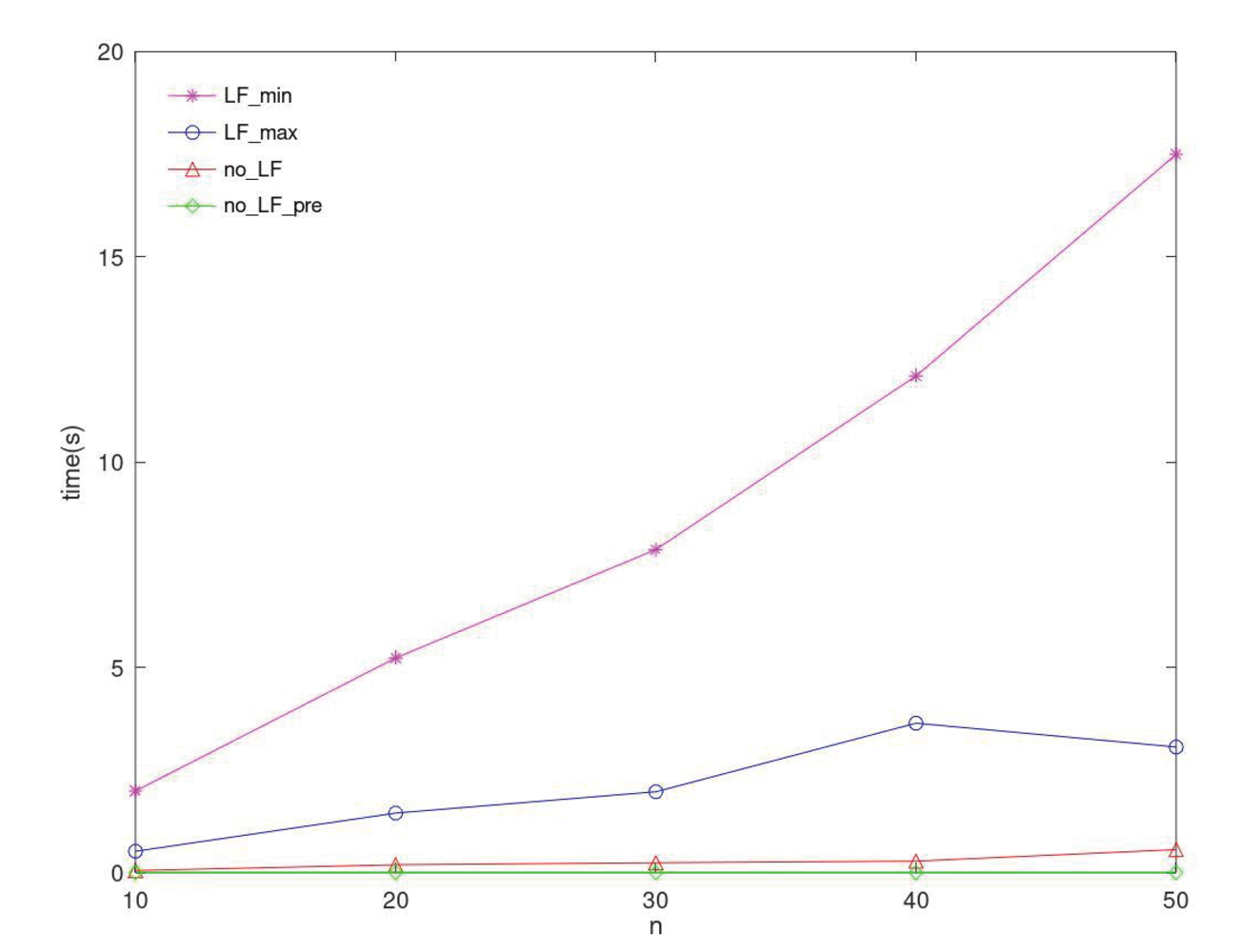}
    \subcaption{Time to find a stable model}\label{fig:4}
  \end{minipage}
& \hspace{2em} &
  \begin{minipage}[b]{0.4\linewidth}\centering
    \includegraphics[keepaspectratio,scale=0.155]{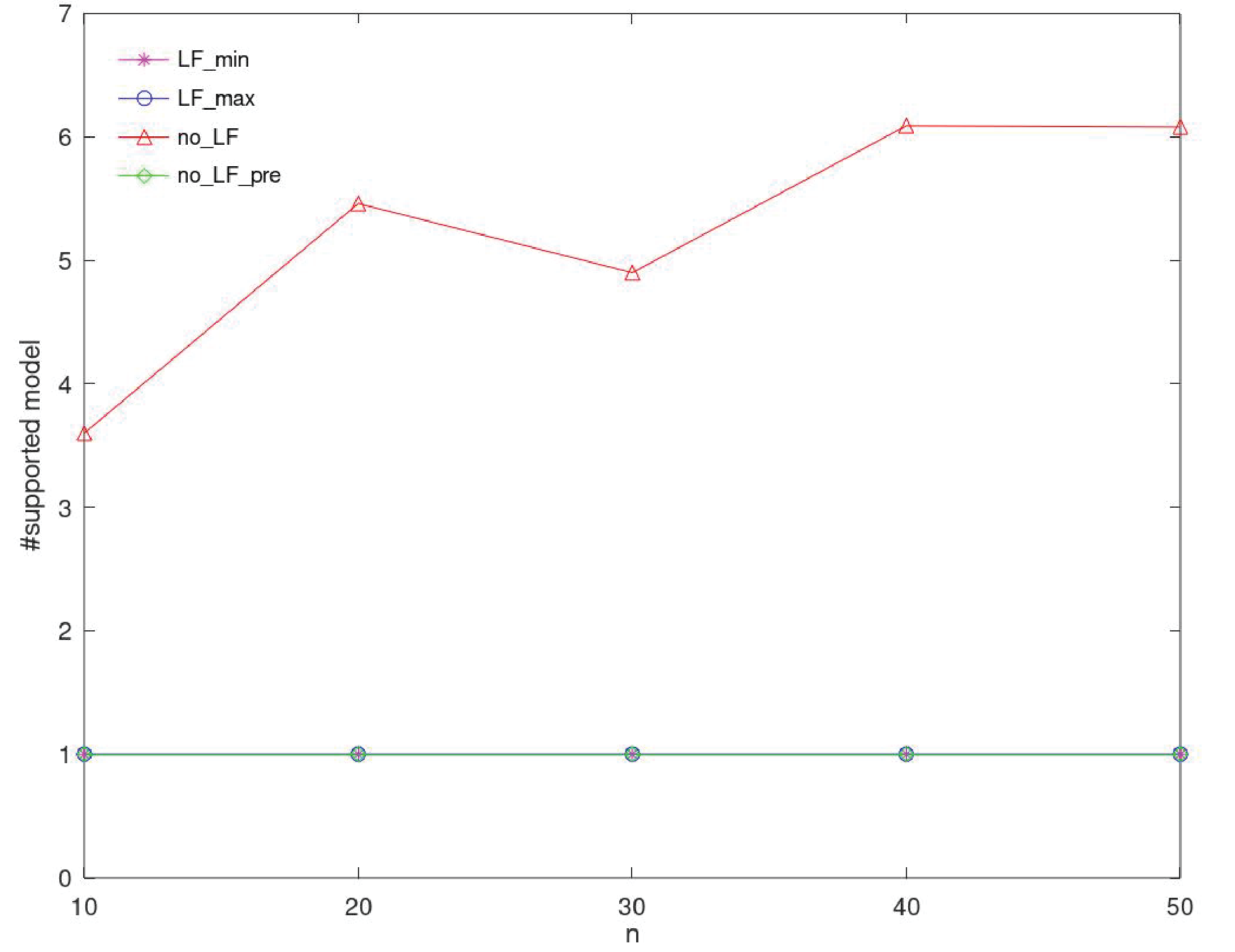}
    \subcaption{\#computed models}\label{fig:5}
  \end{minipage}
\end{tabular}
\caption{The effect of LF heuristics and precomputation on program $P_{4\_n}$}
\label{graph:LF}
\end{figure}

We  next  compare the  effectiveness  of  LF  heuristics and  that  of
precomputation   under  another   solution  constraint.    For  $n   =
10,\ldots,50$, we repeatedly run Algorithm 1 using $L_{SU+\widehat{\matr{c}}+LF}$
with  $max\_try  = 10,  max\_itr  =  100$  on matricized  $\matr{P}_{4\_n}  =
(\matr{C}_{4\_n},\matr{D}_{4\_n})$ (and  no constraint matrix) to  compute supported
(stable) models.  Coefficients  in $L_{SU+\widehat{\matr{c}}+LF}$ are set  to $\ell_2 =
0.1, \ell_3 = 0, \ell_4 = 1$.
To  be  more precise,  for  each  $n$  and  each case  of  $LF_{max}$,
$LF_{min}$, precomputation  (without $\{LF_{max},LF_{min} \}$)  and no
$\{  LF_{max},  LF_{min},  \mbox{precomputation}   \}$,  we  run   Algorithm 1  at most 100  trials to measure  time to find  a stable
model and count the number of supported models computed till then.  We
repeat  this computation  ten times  and take  the average  and obtain
graphs in Fig.~\ref{graph:LF}.

% ave. time, #trial, #stable over ten runs
%--(another solution) constraint used -> each trial seeks for a new supported model
% n = 10; max_try = 10;  max_itr = 100;  max_fp = 100; l2=l3=0.1, l5=1.0

In Fig.~\ref{graph:LF}, no\_LF means no  use of $\{ LF_{max}, LF_{min}
\}$  heuristics.    Also  no\_LF\_pre  means  no\_LF   is  applied  to
precomputed $P_{4\_n}$\footnote{
Precomputation takes 0.006s and removes only one stable false atom. So
precomputation is not helpful in the current case.
}.

% precomp_time = 6.0759e-03
%  D:(52x103)->(51x102) 5356:5202
%  nQ:(103x104)->(102x102) 10712:10404
%  \widehat{\matr{C}}:(0x0)->(0x0) 0:0
% |asp_false_atoms|/|total atoms| = 1/52: (just one false atom)/(n+2 atoms)
%
We can see from graph (a) in Fig.~\ref{graph:LF} that computation time
is $LF_{min}  > LF_{max} > \mbox{no\_LF}  > \mbox{no\_LF\_pre}$.  This
means that using LF heuristics is not necessarily a good policy.  They
might cause extra computation to reach the same model.
Concerning the number of non-stable models computed redundantly, graph
(b) in  Fig.~\ref{graph:LF} tells us that  $LF_{min}$ allows computing
redundant  non-stable  models but  the  rest,  $LF_{max}$, no\_LF  and
no\_LF\_pre,  return  a  stable   model  without  computing  redundant
non-stable models.   This shows first that  $LF_{max}$ works correctly
to suppress the  computation of non-stable models and  second that the
$LF_{min}$ heuristics works adversely, i.e.,guiding the search process
away from the stable model.  This somewhat unexpected result indicates
the need of (empirical) choice of LF heuristics.\\

%program-19
%-----------------
%%[ Loopy program P_loop3_n+k ]:
%%[ Loopy program P_loop3_n+k ]:
% n:even  n+k atoms {a0,a1..an,a(n+1)..a(n+k)}
%   a0:- a1,..,an                        \
%   a(2i-1):- a0 v a(2i)   for i=1..n/2  |  P_loop1_n
%   a(2i):- a0 v a(2i-1)   for i=1..n/2  /
%   a(n+1):- a(n+1)
%   ..
%   a(n+k):- a(n+k)
%  *a0:- ~a(n+1)&..&~a(n+k)
% n:even  n+k atoms {a0,a1..an,a(n+1)..a(n+k)}
%%(Comment):
%% <>  no j-iteration performed, ini. thresholding gives a solution in all cases
%% ($) |asp_false_atoms|/|total atoms| = 5000/10001
%%     precomp elapsed_time = 0.000005
%%     D:(10001x15002)->(5001x10002) 150035002:50020002     marix size -> 1/3
%%     nQ:(15002x20002)->(10002x10002) 300070004:100040004
%%     \widehat{\matr{C}}:(0x0)->(0x10002) 0:0
%% ----------------------------------------------------------------------
%% <>  Our_ASP  >  Clingo(version 5.6.2)   by "> cd clingo; clingo  P_loop3_nk.lp P_loop3_nk_fact.lp"
%%     (0.042s)    (0.027s)  at n = k = 5*10^3

%% [ Ex-19-precomp ]:  precomputation gets closer to Clingo (clingo version 5.6.2)
% [ Clingo ]: find ONE stable model of P_loop3_n+k            FMV WSL2, 2023/02/27
% [ Our_ASP ]: find ONE stable model of P_loop3_n+k           FMV WSL2, 2023/02/27
% ave over ten runs
% no_LF + (precomp) in asp_c.m, max_try=10, max_itr=100, max_fp=100, l2=..=l5=0.1

\begin{figure}[htbp]
\begin{tabular}{ccc}
  \begin{minipage}[b]{0.4\linewidth}\centering
\begin{eqnarray*}
P_{5\_nk} & = &
     \begin{array}{ll}
     \left\{\;
          \begin{array}{lll}
            a_0       & \leftarrow\,  &  a_1 \wedge\ldots\wedge a_n \\
            a_{0}     & \leftarrow\,  &  \neg a_{n+1}\wedge\ldots\wedge \neg a_{n+k}  \\
            \ldots  \\
            a_{2i-1}   & \leftarrow\,  & a_0 \vee a_{2i}    \\ % \hspace{2em} \mbox{for  $i:1 \leq i \leq n/2$} \\
            a_{2i}    & \leftarrow\,  &  a_0 \vee a_{2i-1}  \\ % \hspace{1em}  \mbox{for  $i:1 \leq i \leq n/2$} \\
            \ldots  \\
            a_{n+1}    & \leftarrow\, &  a_{n+1} \\
            \ldots  \\
            a_{n+k}    & \leftarrow\, &  a_{n+k} \\
          \end{array}
     \right.
     \end{array}  \label{prog:loop2}
\end{eqnarray*}
   \subcaption{A non-tight program $P_{5\_nk}$}\label{fig:6}
   \end{minipage}
& \hspace{1em} &
    \begin{minipage}[b]{0.4\linewidth}\centering
    \includegraphics[keepaspectratio,scale=0.29]{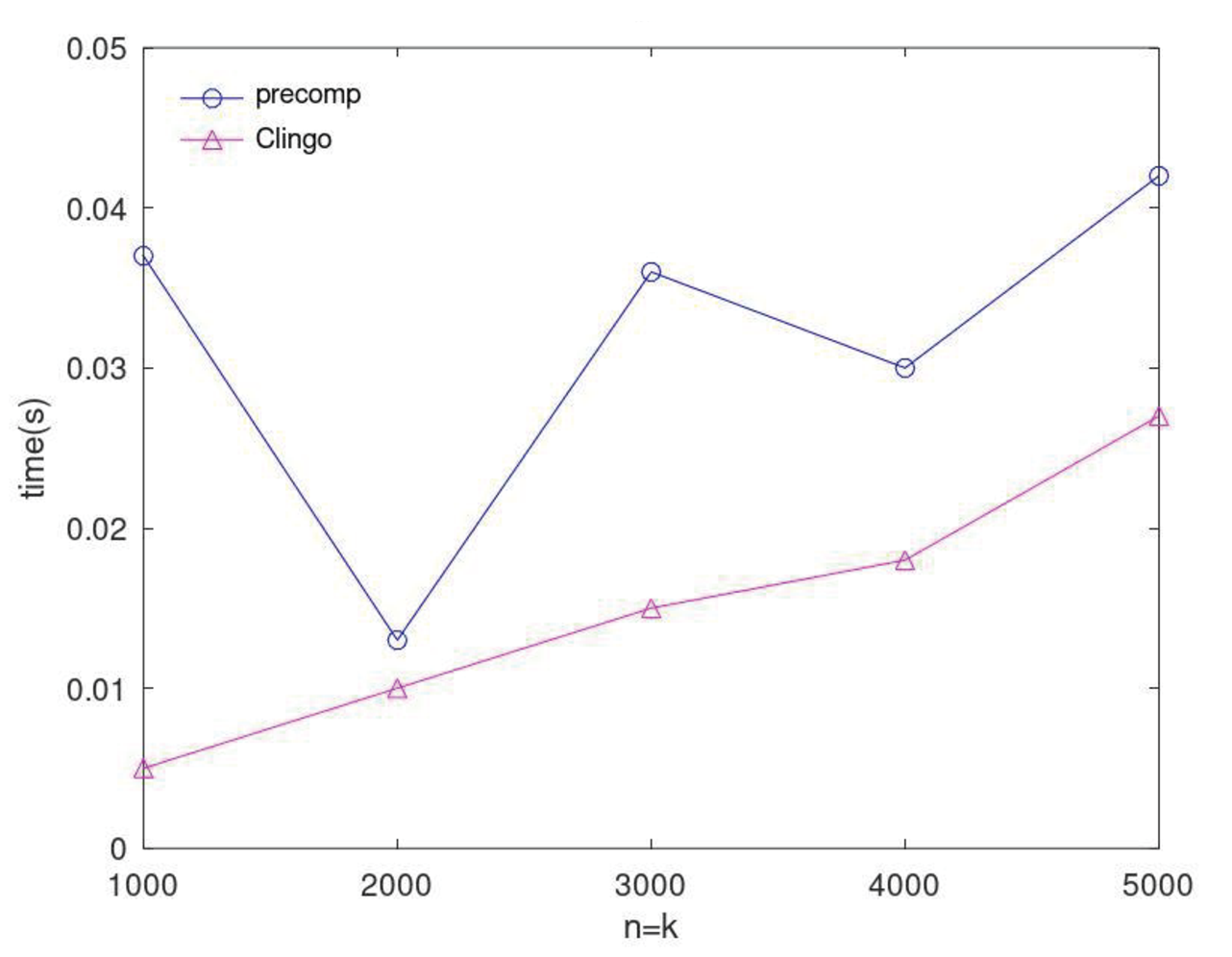}
    \subcaption{Scalability of precomputation w.r.t. $P_{5\_nk}$}\label{fig:7}
   \end{minipage}
\end{tabular}
\caption{Precomputation applied to program $P_{5\_nk}$}
\label{graph:P5}
\end{figure}

Finally to examine  the effectiveness of precomputation  more precisely, we
apply precomputation to  a more complex program $P_{5\_nk}$.   It is a
modification  of  $P_{4\_n}$ by  adding  self-loops  of $k$  atoms  as
illustrated  by (a)  in Fig.~\ref{graph:P5}.   The addition  of self-loop
causes the choice of $a_{n+j}$ ($1 \leq j \leq k$) being true or being
false in the search process.  $P_{5\_nk}$ has $(2^{n/2}-1)(2^{k}-1)+1$
supported models but has just  one stable model $\{ a_0,a_1,\ldots,a_n
\}$.

We compute a stable model by  running Algorithm 1 on precomputed
$P_{5\_nk}$ without  using LF heuristics up  to $n = k  = 5000$.  When
precomputation  is applied  to $P_{5\_nk}$  where $n  = k  = 5000$,  it
detects  $5000$  false  stable   atoms  and  downsizes  the matrices in
$\matr{P}_{5\_nk}  =  (\matr{C}_{5\_nk}, \matr{D}_{5\_nk})$ from  $\matr{D}_{5\_nk}(10001  \times
15002)$ to  $\matr{D}'_{5\_nk}(5001 \times 10002)$ and  from $\matr{C}_{5\_nk}(15002
\times 20002)$  to $\matr{C}'_{5\_nk}(10002 \times 10002)$.  Thus precomputed
$\matr{P}'_{5\_nk} =  (\matr{C}'_{5\_nk}, \matr{D}'_{5\_nk})$  is downsized  to 1/3  of the
original $P_{5\_nk}$.

We run Algorithm 1 on  $P'_{5\_nk}$ with $\ell_2 = \ell_3 = 0.1$
and $max\_try = 10$, $max\_itr=100$ at most 100 trials to measure time
to find a  stable model ten times for each  $n = 1000,\ldots,5000$ and
take  the average.
% Look at
% if 0  %program-19   in ./Octave/Costminization/ASP/test_asp_c.m
%
At the  same time, we also  run \textit{clingo} (version 5.6.2)  on $P_{5\_nk}$
and similarly measure  time.
% Look at
% if 0  ~sato/clingo/P_loop3_nk_fact.lp and P_loop3_nk.lp
%
Graph (b) in Fig.~\ref{graph:P5} is the  result.  It shows that as far
as computing a stable model  of $P_{5\_nk}$ is concerned, our approach
comes  close to  \textit{clingo}.   However, this  is due  to  a very  specific
situation   that   precomputation   removes  all   false   atoms   $\{
a_{n+1},\ldots,a_{n+k} \}$ in the stable model of $P_{5\_nk}$ and  Algorithm  1  run on  the  precomputed  $\matr{P}'_{5\_nk} =  (\matr{C}'_{5\_nk},
\matr{D}'_{5\_nk})$ detects  the stable model only  by thresholding $\mbf{s}$
before starting  any update  of $\mbf{s}$.  So  what graph  (b) really
suggests seems to be the importance  of optimization of a program like
precomputation, which is to be developed further in our approach.

%% ($) |asp_false_atoms|/|total atoms| = 5000/10001
%%     precomp elapsed_time = 0.000005
%%     D:(10001x15002)->(5001x10002) 150035002:50020002     marix size -> 1/3
%%     nQ:(15002x20002)->(10002x10002) 300070004:100040004
%%     \widehat{\matr{C}}:(0x0)->(0x10002) 0:0

% ave over ten runs
% no_LF + (precomp) in asp_c.m, max_try=10, max_itr=100, max_fp=100, l2=..=l5=0.1

\section{Related work}\label{related}
The   most    closely   related    work   is    \cite{Aspis2020}   and
\cite{Takemura2022}.    As  mentioned   in  Section~\ref{intro},   our
approach differs  from them  in three  points: (1)  theoretically, the
exclusion of non-stable models by loop formulas, (2) syntactically, no
restriction on acceptable programs  and (3) practically, incorporation
of constraints.  Concerning  performance, they happen to  use the same
$N$-negative loops  program which  consists of $N$  copies (alphabetic
variants) of a program $\{a\leftarrow  \neg b, b\leftarrow \neg a \}$.
According  to  \cite{Aspis2020},  the   success  rate  w.r.t.  $N$  of
returning a supported model goes from  one initially to almost zero at
$N =  64$ in  \cite{Aspis2020} while it  keeps one till  $N =  100$ in
\cite{Takemura2022}.  We tested the same program with $max\_try = 20$,
$max\_itr = 100$ and observed that  the success rate keeps one till $N
= 10000$.

Although  our  approach  is   non-probabilistic,  i.e., purely  linear
algebraic, there are probabilistic  differentiable approaches for ASP.
Differentiable ASP/SAT \cite{Nickles18a}  iteratively samples a stable
model by an ASP solver a la ASSAT \cite{Lin04}. The solver decides the
next  decision literal  based on  the derivatives  of a  cost function
which  is  the MSE  between  the  target probabilities  and  predicted
probabilities computed  from the sampled stable  models via parameters
associated with ``parameter atoms'' in a program.

NeurASP  \cite{Yang20} uses  an  ASP solver  to  obtain stable  models
including  ``neural atoms''  for a  program.  They  are associated  with
probabilities  learned  by deep  learning  and  the likelihood  of  an
observation  (a set  of ASP  constraints) is  computed from  them. The
whole learning  is carried  out by  backpropagating the  likelihood to
neural atoms to parameters in a neural network.

Similarly to  NeurASP, SLASH \cite{Skryagin22,Skryagin23}  uses an ASP  solver to
compute stable  models for a program  containing ``neural probabilistic
predicates''. Their probabilities are dealt with by neural networks and
probabilistic  circuits. The  latter makes  it possible  to compute  a
joint distribution of the class category and data.
Both NeurASP and SLASH are examples of symbolic ASP solver-based neuro-symbolic systems,
where they include a neural frontend to process the perception part of the problem,
and a symbolic backend which typically is the ASP solver.
Therefore, the neural frontend does not need to be involved in the computational details and problems associated with  computing stable models (Section \ref{stable}).

Independently of ASP solver-based approaches mentioned above, Sato and
Kojima proposed a differentiable approach to sampling supported models
of   (non-propositional)    probabilistic   normal    logic   programs
\cite{Sato19,Sato20}.  They encode programs  by matrices and formulate
the problem  of sampling  supported models  as repeatedly  computing a
fixedpoint of some differentiable  equations. They solve the equations
in vector spaces by minimizing a non-negative cost function defined by
Frobenius norm.
More recently, Takemura and Inoue \cite{Takemura2024} proposed a neuro-symbolic learning pipeline for distant supervision tasks,
which leverages differentiable computation of supported models.
Similarly to this work, they encode normal logic programs into matrices and define a differentiable loss function 
which is based on the supported model semantics.

As for the non-differentiable linear-algebraic approaches to logic programming,
Nguyen et al.\ adopted  matrix encoding for propositional  normal logic programs
based  on \cite{Sakama17}  and proposed  to compute  stable models  in
vector   spaces  by   a   generate-and-test   approach  using   sparse
representation \cite{Nguyen22}.

\subsection{Connection to neural network computation}\label{NN}
At this  point, it is quite  interesting to see the  connection of our
approach to neural network computation.  In (\ref{eq:jsu}), we compute
$\matr{M}$ and  $\mbf{d} = \matr{DM}$.   We point out  that the computation  of this
$\mbf{d}$ is nothing  but the output of a forward  process of a single
layer  ReLU network  from  an input  vector  $\mbf{s}$.  Consider  the
computation of  $\matr{M} = (\matr{1}-\rmin_1(\matr{C}(\matr{1}-\mbf{s}^{\delta})))$.   We rewrite
this   using  $1-{\rm   min}(x,1)   =  {\rm   ReLU}(1-x)$   to
\begin{eqnarray*}
  \matr{M}  & = & \matr{1}-\rmin_1(\matr{C}^{pos}(\matr{1}-\mbf{s})+\matr{C}^{neg}\mbf{s}) \\
     & = & {\rm  ReLU}(\matr{W}\mbf{s}  + \mbf{b}) \\
     &   & \mbox{where}\;\;
           \matr{C} = [\matr{C}^{pos}\, \matr{C}^{neg}],\, \matr{W} = \matr{C}^{pos}-\matr{C}^{neg},\,
           \mbf{b} = \matr{1}-\matr{C}^{pos}\mbf{1}
\end{eqnarray*}
\noindent
So $\matr{M}$  is the output of  a ReLU network  having a weight matrix  $\matr{W} =
\matr{C}^{pos}-\matr{C}^{neg}$ and a bias vector $\mbf{b} = \matr{1}-\matr{C}^{pos}\mbf{1}$.  Then
${\rm  min}_1(\mbf{d})  =  {\rm min}_1(\matr{DM})  =  {\rm  min}_1(\matr{D}\cdot{\rm
  ReLU}(\matr{W}\mbf{s} + \mbf{b}))$  is the output of a ReLU  network with a
single  hidden layer  and a  linear  output layer  represented by  $\matr{D}$
having ${\rm min}_1(\cdot)$ as activation function.

Also when we compute a supported model $\mbf{s}$, we minimize $L_{SU+\widehat{\matr{c}}}$
(\ref{eq:jsu})  which contains  an MSE  error term  $L_{sq} =  \| {\rm
  min}_1(\mbf{d})    -    \mbf{s}   \|^{2}$    using    $J_{a_{SU+\widehat{\matr{c}}}}$
(\ref{eq:jasuc}).  This  is precisely  back propagation  from learning
data $\mbf{s}$.

Thus we may  say that our approach is an  integration of ASP semantics
and neural  computation and  provides a  neuro-symbolic \cite{Sarker2021}
way of  ASP computation.  Nonetheless,  there is a big  difference.  In
standard neural network  architecture, a weight matrix $\matr{W}$  and a bias
vector   $\mbf{b}$   are  independent.   In   our   setting,  they   are
interdependent and they faithfully reflect  the logical structure of a
program.

\section{Conclusion}\label{conclusion}
We  proposed  an  end-to-end  approach  for  computing  stable  models
satisfying given constraints.  We matricized a program and constraints
and formulated stable  model computation as a  minimization problem in
vector spaces  of a  non-negative cost function.   We obtain  a stable
model satisfying constraints  as a root the cost  function by Newton's
method.

By incorporating all loop formula constraints introduced in Lin-Zhao's
theorem \cite{Lin04}  into the cost  function to be minimized,  we can
prevent redundant  computation of  non-stable models,  at the  cost of
processing  exponentially many  loop formulas.   Hence, we  introduced
precomputation which downsizes a program while preserving stable model
semantics and also two heuristics that selectively use loop formulas.
Then we confirmed the effectiveness of  our approach including
precomputation and loop formula heuristics by simple examples.

Future work could focus on improving the integration of neural networks with this proposed end-to-end approach to tackle neuro-symbolic benchmark tasks that require both perception and reasoning.
We also aim to improve the optimization techniques, such as precomputation, to enhance efficiency and scalability.

\begin{acks}
This work is supported by JSPS KAKENHI Grant Numbers JP21H04905, JP25K03190 and JST CREST Grant Number JPMJCR22D3.
\end{acks}

\begin{appendix}

\section{Proofs}

\subsection{Proposition 2}\label{sec:appdx_prop2}

Recall that we write $\matr{C} \in \{0,1\}^{m \times 2n}$ as $\matr{C} = [\matr{C}^{pos}\; \matr{C}^{neg}]$ where
$\matr{C}^{pos}  \in \{0,1\}^{m  \times n}$  (resp.  $\matr{C}^{neg}  \in \{0,1\}^{m
  \times  n}$)  is the  left  half  (resp.   the  right half)  of  $\matr{C}$
representing the  positive literals (resp. negative  literals) of each
rule body in $\matr{C}$.
Let $\matr{M}^{neg} = \matr{1}-\rmin_1(\matr{C}^{neg}\mbf{s}_{I})$, and let $\matr{M}^{pos} = \matr{M}^{neg}\odot  (\matr{1}  - \rmin_1(\matr{C}^{pos}(\matr{1}-\mbf{s}_{I})))$.

We  prove $\mbf{d}_I  = \mbf{d}_I^{+}$  first.  
Recall  that a rule $r_j^{+}$  in $P^{I}$  is created  by removing  negative literals true in $I$ from the body of $r_j$  in $P$.  
So for any $a_i \in {\cal A}$, it is immediate that $a_i$ has a rule $r_j \in P$ whose body is true in $I$ iff  $a_i$ has the rule $r_j^{+} \in  P^{I}$ whose body is true in $I$.  
Thus $\mbf{d}_I(i) = \mbf{d}_I^{+}(i)$ for every $i$ ($1 \leq i \leq n$), and hence $\mbf{d}_I = \mbf{d}_I^{+}$.  Consequently, we  have $\|  \mbf{s}_I -  {\rm min}_1(\mbf{d}_I)  \|_2 =  0$ iff  $\|\mbf{s}_I -  {\rm min}_1(\mbf{d}_I^{+})  \|_2 =  0$.

$I \models \mbox{comp}(P^I)$ iff $\| \mbf{s}_I  - {\rm min}_1(\mbf{d}_I^{+}) \|_2 =  0$ is  proved similarly  to Proposition~\ref{prop1}.
Firstly, suppose that \(I\) is a supported model of \(P^I\). 
By definition, for each \(a_i\) that is true in \(I\), there is at least one rule body in \(P^I\) that is true in \(I\), i.e., \(I \models a_i\) and \(I \models G_{j1} \vee \cdots \vee G_{js}\).
Since \(\matr{M}^{pos}\) denotes the truth values of the rule bodies in \(P^I\) evaluated by \(I\), \(\matr{M}^{pos}(j) = 1 \; (1 \leq j \leq m)\) if \(r^+_j\) is contained in \(P^I\) and its body is true in \(I\), otherwise \(\matr{M}^{pos}(j)=0\).
Let \(\mbf{d}_I^{+}=\matr{D}\matr{M}^{pos}\) where \(\mbf{d}_I^{+}(i) \geq 1\) if \(a_i\) is true in \(I\), then we have \(\mbf{s}_I(i)={\rm min}_1(\mbf{d}_I^{+})(i)\).
Since \(i\) is arbitrary, we conclude \(\mbf{s}_I=\mbf{d}_I^{+}\), and that \(\| \mbf{s}_I  - {\rm min}_1(\mbf{d}_I^{+}) \|_2 =  0\).
Secondly, suppose that \(\| \mbf{s}_I  - {\rm min}_1(\mbf{d}_I^{+}) \|_2 =  0\).
Since we have \({\rm min}_1(\mbf{d}_I^{+})(i) = 1\) if any of the rule bodies for \(a_i\) is true in \(I\), \(I \models G_{j1} \vee \cdots \vee G_{js}\) which we denote \(I \models {\rm iff}(a_i)\).
Because \(i\) is arbitrary, and for all atoms that are true in \(I\) the above condition holds, we conclude \(I \models {\rm comp}(P^I)\).

Here, we proved $\|  \mbf{s}_I -  {\rm min}_1(\mbf{d}_I)  \|_2 =  0$ iff  $\|\mbf{s}_I -  {\rm min}_1(\mbf{d}_I^{+})  \|_2 =  0$, and $I \models \mbox{comp}(P^I)$ iff $\| \mbf{s}_I  - {\rm min}_1(\mbf{d}_I^{+}) \|_2 =  0$.
From Proposition~\ref{prop1}, we have that $I \models \mbox{comp}(P)$ iff $\| \mbf{s}_I  - {\rm min}_1(\mbf{d}_I) \|_2 =  0$. 
Hence, 
$I \models {\rm comp}(P)$,
$\| \mbf{s}_I - {\rm min}_1(\mbf{d}_I) \|_2 = 0$,
$\| \mbf{s}_I - {\rm min}_1(\mbf{d}_I^{+}) \|_2 = 0$ and
$I \models {\rm comp}(P^{I})$
are all equivalent. \hspace{1em} Q.E.D.\\
% %     Na = Q2*a;                       % if row in Q2 is zero vector -> Na=0 -> Ma=1
% %     Ma = 1 - (Na>=1);                % negative bodies evaluated by a
% %     GLa = Ma.*(1 - min(Q1*(1-a),1)); % Ma = indicator of survived bodies in nQ
% %     DGLa = (D*GLa)>=1;               % if a has no disjunct in D -> a is false
% %     error_fp = sum(abs(a - DGLa));   % l1_norm error

\section{Derivations}

\subsection{A note on the matrix-matrix dot product notation}\label{sec:appdx_matrix_notation}
\subsubsection{\texorpdfstring{$\matr{A} \bullet (\matr{B}\odot\matr{C}) = (\matr{B}\bullet\matr{A}) \odot \matr{C}$}{}}
In various parts of this paper, we use the notation of the dot product of matrices $(\matr{A} \bullet \matr{B}) = \sum_{i,j}\matr{A}(i,j)\matr{B}(i,j)$.
This is essentially an element-wise multiplication (the Hadamard product) followed by the summation of matrix elements.
Since the Hadamard product is associative and commutative, it follows that $(\matr{A} \odot (\matr{B}\odot\matr{C})) = ((\matr{B}\odot\matr{A}) \odot \matr{C})$ holds. The summation is also associative and commutative; thus $(\matr{A} \bullet (\matr{B}\odot\matr{C})) = ((\matr{B}\bullet\matr{A}) \odot \matr{C})$ holds.

\subsubsection{\texorpdfstring{$\matr{A} \bullet (\matr{B}\matr{C}) = (\matr{B}^T\matr{A}) \bullet \matr{C}$}{}}
Using the aforementioned \(\bullet\) notation, we have $(\matr{A} \bullet (\matr{B}\matr{C})) = \sum_{i,j} \matr{A}(i,j)(BC)(i,j) = ((\matr{B}^T\matr{A}) \bullet \matr{C}) = \sum_{i,j}(B^T\matr{A})(i,j)\matr{C}(i,j)$.
Without the summation, this will not hold because the matrix multiplication is not commutative in general.
Consider a square \(N \times N\) matrix \(\matr{A}, \matr{B}\) and \(\matr{C}\), then we have:
\begin{eqnarray*}
\matr{A} \odot (\matr{BC})\; & = &
  \begin{bmatrix}
  \matr{A}(1,1)\sum_{i=j}^{N}\matr{B}(1,j)\matr{C}(i,1) & \cdots & \matr{A}(1,N)\sum_{i=j}^{N}\matr{B}(1,j)\matr{C}(i,N) \\
  \vdots & & \vdots \\
  \matr{A}(N,1)\sum_{:i=j}^{N}\matr{B}(N,j)\matr{C}(i,1) & \cdots & \underline{\matr{A}(N,N)\sum_{i=j}^{N}\matr{B}(N,j)\matr{C}(i,N)} \\
  \end{bmatrix}
  \\
  \textrm{and}
  \\
  (\matr{B}^T \matr{A}) \odot \matr{C}\; & = &
  \begin{bmatrix}
  \matr{C}(1,1)\sum_{i}^{N}\matr{A}(i,1)\matr{B}(i,1) & \cdots & \matr{C}(1,N)\sum_{i}^{N}\matr{A}(i,N)\matr{B}(i,1) \\
  \vdots & & \vdots \\
  \matr{C}(N,1)\sum_{i}^{N}\matr{A}(i,1)\matr{B}(i,N) & \cdots & \matr{C}(N,N)\sum_{i}^{N}\matr{A}(i,N)\matr{B}(i,N) \\
  \end{bmatrix}
\end{eqnarray*}
Then, observe that an element in one matrix appears in the expanded elements in the columns of the other matrix.
For example, consider the \(\matr{A}(N,N)\) term in \(\matr{A} \odot (\matr{BC})\) (bottom-right), and notice that the term containing \(\matr{A}(N,N)\) appears only in the rightmost column, and that they appear in the last elements after expansion, i.e.:
\begin{eqnarray*}
\matr{C}(1,N)\sum_{i}^{N}\matr{A}(i,N)\matr{B}(i,1) &=& \matr{C}(1,N)\Bigl( \matr{A}(1,N)\matr{B}(1,1) + \cdots + \matr{A}(N,N)\matr{B}(N,1)\Bigr) \\
                               &=& \matr{A}(1,N)\matr{B}(1,1)\matr{C}(1,N) + \cdots + \underline{\matr{A}(N,N)\matr{B}(N,1)\matr{C}(1,N)} \\
\matr{C}(N,N)\sum_{i}^{N}\matr{A}(i,N)\matr{B}(i,N) &=& \matr{C}(N,N)\Bigl( \matr{A}(1,N)\matr{B}(1,N) + \cdots + \matr{A}(N,N)\matr{B}(N,N)\Bigr) \\
                               &=& \matr{A}(1,N)\matr{B}(1,N)\matr{C}(N,N) + \cdots + \underline{\matr{A}(N,N)\matr{B}(N,N)\matr{C}(N,N)} \\
\end{eqnarray*}
Since the summation operation in \(\bullet\) is commutative, one can collect the underlined parts to the summation, \(\matr{A}(N,N)\sum_{i=j}^{N}\matr{B}(N,j)\matr{C}(i,N)\), which is the bottom-right element in \(\matr{A} \odot (\matr{BC})\).
One can show that this applies to all other elements of the matrix, e.g., the \(\matr{C}(N,N)\) term in \((\matr{B}^T \matr{A}) \odot \matr{C}\) appears in the last elements in the rightmost column of \(\matr{A} \odot (\matr{BC})\). 
Therefore, $(\matr{A}   \bullet  (\matr{B}\matr{C}))  = ((\matr{B}^T\matr{A}) \bullet \matr{C})$ holds. 

\subsection{\texorpdfstring{\(J_{a_{SU}}\)}{}: Jacobian for supported model computation (Section \ref{supported})}\label{sec:appdx_supported}
Let $\matr{P} = (\matr{C},\matr{D})$ be the matricized program and write $\matr{C} = [\matr{C}^{pos}\,\matr{C}^{neg}]$.
Introduce $\matr{N}$, $\matr{M}$, $\mbf{d}$, $\matr{E}$, $\matr{F}$  and compute $L_{SU}$  by
\begin{eqnarray*}
\begin{array}{lclll}
      \matr{N} & =\;\; & \matr{C}(\matr{1}-\mbf{s}^{\delta}) = \matr{C}^{pos}(\matr{1}-\mbf{s})+\matr{C}^{neg}\mbf{s}
                                              & & \mbox{: (continuous) counts of false literals in the rule bodies} \\
      \matr{M} & =\;\; & \matr{1} - {\rm min}_1(\matr{N})              & & \mbox{: (continuous) truth values of the rule bodies}  \\
\mbf{d} & =\;\; & \matr{\matr{DM}}                              & & \mbox{: (continuous) counts of true disjuncts for each atom}  \\
      \matr{E} & =\;\; & {\rm min}_1(\mbf{d}) - \mbf{s}  & & \mbox{: error between the estimated truth values of atoms and $\mbf{s}$}  \\
      \matr{F} & =\;\; & \mbf{s} \odot (\matr{1} - \mbf{s}) & & \mbox{: (continuous) 0 iif $\matr{s}$ is binary } \\
    L_{sq} & =\;\; & (\matr{E} \bullet \matr{E}) \\
   L_{nrm} & =\;\; & (\matr{F} \bullet \matr{F}) \\
    L_{SU} & =\;\; & 0.5 \cdot (L_{sq} + \ell_2 \cdot L_{nrm}).
\end{array}
\end{eqnarray*}\\

We  first compute
$\displaystyle{ \frac{\partial  L_{sq}}{\partial  {s}_p} }$  where
${s}_p = \mbf{s}(p)$ ($1 \leq p \leq n$).
\begin{eqnarray*}
 \frac{\partial \matr{M}}{\partial s_p}
   & = & -[\matr{N} \leq 1] \odot \bigl((\matr{C}^{neg}-\matr{C}^{pos}) \mbf{I}_p \bigr) = [\matr{N} \leq  1]\odot \bigl( (\matr{C}^{pos}-\matr{C}^{neg}) \mbf{I}_p \bigr) \\
 \frac{\partial L_{sq}}{\partial s_p}
  & = & \Bigl( \matr{E}\,\bullet\; [\matr{DM}\leq 1]\odot \Bigl( D \left( \frac{\partial M}{\partial s_p} \right) \Bigr)  - \mbf{I}_p \Bigr)\\
  & = & \Bigl( \matr{E}\,\bullet\; [\matr{DM}\leq 1]\odot (\matr{D}( [N\leq 1]\odot ((\matr{C}^{pos}-\matr{C}^{neg})\mbf{I}_p) )) - \mbf{I}_p \Bigr)\\
% & = & \Bigl( \matr{E}\,\bullet\; [\matr{DM}\leq 1]\odot (\matr{D}( [N\leq 1]\odot ((\matr{C}^{pos}-\matr{C}^{neg})\mbf{I}_p) )) \Bigr) - ( E \;\bullet\; \mbf{I}_p) \\
  & = & \Bigl( \matr{D}^T([\matr{D M} \leq 1] \odot E) \;\bullet\; [N\leq 1] \odot (((\matr{C}^{pos}-\matr{C}^{neg})\mbf{I}_p) \Bigr)   -  ( E \;\bullet\; \mbf{I}_p )\\
% & = & \Bigl( (\matr{C}^{pos}-\matr{C}^{neg})^T  ([\matr{N} \leq 1] \odot (\matr{D}^T ([\matr{DM} \leq 1] \odot E))) \;\bullet\; \mbf{I}_p \Bigr)  - ( E \;\bullet\; \mbf{I}_p )\\
  & = & \Bigl( (\matr{C}^{pos}-\matr{C}^{neg})^T  ([\matr{N} \leq 1] \odot (\matr{D}^T ([\matr{DM} \leq 1] \odot E))) - E  \;\bullet\; \mbf{I}_p \Bigr)
\end{eqnarray*}
\noindent
Since $p$ is arbitrary, we have
${\displaystyle \frac{\partial L_{sq}}{\partial \mbf{s}} = (\matr{C}^{pos}-\matr{C}^{neg})^T ([\matr{N} \leq 1] \odot (\matr{D}^T ([\matr{DM} \leq 1] \odot \matr{E}))) - \matr{E} }$.
\noindent\\
Next we compute  ${\displaystyle \frac{\partial L_{nrm}}{\partial s_p}}$:
\begin{eqnarray*}
\frac{\partial \matr{F}}{\partial s_p}
  & = & \left( \frac{\partial \mbf{s}}{\partial s_p} \right) \odot (\matr{1} - \mbf{s})
          +   \mbf{s} \odot \left( \frac{\partial (\matr{1}-\mbf{s})}{\partial s_p} \right) \\
  & = & ( \mbf{I}_p \odot (\matr{1} - \mbf{s}) ) - (\mbf{s} \odot \mbf{I}_p)  =   (\matr{1} - 2\mbf{s}) \odot \mbf{I}_p \\
\frac{\partial L_{nrm}}{\partial s_p}
  & = & \bigl( \matr{F} \;\bullet\; \left( \frac{\partial \matr{F}}{\partial s_p} \right) \bigr)\\
  & = & ( \matr{F} \;\bullet\; (\matr{1} - 2\mbf{s}) \odot \mbf{I}_p )
    =  ( (\matr{1} - 2\mbf{s}) \odot \matr{F}  \;\bullet\;  \mbf{I}_p ) \\
\end{eqnarray*}
\noindent
Again since $p$ is arbitrary,
we have ${\displaystyle  \frac{\partial L_{nrm}}{\partial \mbf{s}} = (\matr{1} - 2\mbf{s}) \odot \matr{F}}$
and reach
\begin{eqnarray*}
J_{a_{SU}}
 & = & \left( \frac{\partial L_{sq}}{\partial \mbf{s}} \right)
          + \ell_2 \cdot  \left( \frac{\partial L_{nrm}}{\partial \mbf{s}} \right)  \nonumber\\
 & = & (\matr{C}^{pos}-\matr{C}^{neg})^T ([\matr{N} \leq 1] \odot (\matr{D}^T ([\mbf{d} \leq 1] \odot \matr{E}))) - \matr{E}  + \ell_2 \cdot ((\matr{1} - 2\mbf{s}) \odot \matr{F})      \\
 &   & \;\mbox{where}\;\;
          \matr{N} = \matr{C}(\matr{1}-\mbf{s}^{\delta}),\,
          \mbf{d} = \matr{D}(\matr{1} - {\rm min}_1(\matr{N})),\,
          \matr{E} = {\rm min}_1(\mbf{d}) - \mbf{s},\,
          \;\mbox{and}\; \matr{F} = \mbf{s} \odot (\matr{1} - \mbf{s}).   \nonumber
\end{eqnarray*}

\subsection{\texorpdfstring{\(J_{a_{\widehat{\matr{c}}}}\)}{}: Jacobian for constraints (Section \ref{constraints})}\label{sec:appdx_constraints}
Let $\widehat{\matr{C}}  =  [\widehat{\matr{C}}^{pos} \, \widehat{\matr{C}}^{neg}]$ represent the rule bodies of constraints in a binary matrix.
The rest of the derivation is similar to the previous section, namely the derivation of ${\displaystyle \frac{\partial L_{sq}}{\partial \mbf{s}}}$.
% J_constraint: \widehat{\matr{C}} = [Q1_c Q2_c] where \widehat{\matr{C}} represents constraints such as :- a&b&~c
\begin{eqnarray*}
\matr{N}_{\widehat{c}}  & = &  \widehat{\matr{C}}(\matr{1}-\mbf{s}^{\delta}) = \widehat{\matr{C}}^{pos}(\matr{1}- \mbf{s}) + \widehat{\matr{C}}^{neg}\mbf{s}   \nonumber \\
L_{\widehat{\matr{c}}}  & = & (\mbf{1} \bullet (\matr{1} - {\rm min}_1(\matr{N}_{\widehat{c}}))) \hspace{0.5em}\mbox{where $\mbf{1}$ is an all-ones vector} \\
\frac{\partial J_{a_{\widehat{\matr{c}}}}}{\partial \mbf{s}} & = &  -[\matr{N}_{\widehat{c}} \leq 1] \odot \bigl((\widehat{\matr{C}}^{neg}-\widehat{\matr{C}}^{pos}) \mbf{I} \bigr) \\
  & = & (\widehat{\matr{C}}^{pos} - \widehat{\matr{C}}^{neg})^T [\matr{N}_{\widehat{c}} \leq 1] \\
% J_{a_{\widehat{\matr{c}}}} & = & (\widehat{\matr{C}}^{pos} - \widehat{\matr{C}}^{neg})^T [N_c \leq 1] 
\end{eqnarray*}

\subsection{\texorpdfstring{\(J_{a_{LF}}\)}{}: Jacobian for loop formula (Section \ref{loopf})}\label{sec:appdx_loopformula}

Let $\matr{C} = [\matr{C}^{pos}\, \matr{C}^{neg}]$, and let $S_v$, $A_v$ and $\matr{M}$ be computed by (\ref{eq:jLF}).\\
\begin{eqnarray*}
\matr{N}   & = & \matr{C}(\matr{1}-\mbf{s}^{\delta})  \nonumber \\
N_v & = & \matr{S}_v(\matr{1}-\mbf{s}) \nonumber \\
M_v & = & {\rm min}_1(N_v)  \nonumber \\
J_{a_{LF}}
   & = & \frac{\partial L_{LF}}{\partial \mbf{s}}
     =   \sum_{v=1}^{w} - \left( \frac{\partial {\rm min}_1(A_v)}{\partial \mbf{s}} \right) \nonumber \\
   & = & - \sum_{v=1}^{w} [A_v \leq 1]
         \left( \left( \frac{\partial M_v}{\partial \mbf{s}} \right)
               + \matr{S}_v\matr{E}_{sup}\left( \frac{\partial \matr{M}}{\partial \mbf{s}} \right)^T
         \right)                                                                  \nonumber \\
   & = & \sum_{v=1}^{w} [A_v \leq 1]
         \left(   [N_v \leq 1] \matr{S}_v^T + ( ((\matr{S}_v\matr{E}_{sup})\odot[\matr{N} \leq 1]^T)(\matr{C}^{neg}-\matr{C}^{pos})^T
         \right)
\end{eqnarray*}

\section{Encoding the Hamiltonian Cycle Problem}\label{sec:appdx_hamiltonian}
This section describes the encoding and program used in solving the Hamiltonian cycle problem (Section \ref{hamiltonian}).

Firstly, looking at the graph (Fig. \ref{fig:3}), it is evident that there are six vertices.
We use an atom \(H_{i,j}\) to indicate there exists an edge from vertices \(i\) to \(j\) in an HC.
Then there are 36 atoms \(\{ H_{1,1},H_{1,2},\cdots,H_{1,6},H_{2,1},\cdots,H_{6,1},\cdots,H_{6,6} \}\).
We also use an atom \(U_{j,q}\) to indicate that the vertex \(j\) is visited at time \(q\).
Then there are 36 atoms \(\{ U_{1,1},U_{1,2},\cdots,U_{1,6},U_{2,1},\cdots,U_{6,1},\cdots,U_{6,6} \}\).
Thus, in total, there are 72 atoms consisting of \(H_{i,j}\) and \(U_{j,q}\).

\subsection{Encoding (1) (2) (3) into a program}

The condition \(\rm{one\_{of}}\) is encoded as a set of normal rules whose body consists solely of negative literals.
For example, looking at vertex 1, there are 3 outgoing edges to vertices 2, 3 and 4.
Then, we construct rules for (1):
\begin{eqnarray*}
P_{H_{1,j}} & = &
     \begin{array}{ll}
     \left\{\;
          \begin{array}{lll}
            H_{1,2} \leftarrow \neg H_{1,3} \wedge \neg H_{1,4} \\
            H_{1,3} \leftarrow \neg H_{1,2} \wedge \neg H_{1,4} \\
            H_{1,4} \leftarrow \neg H_{1,2} \wedge \neg H_{1,3} \\
          \end{array}
     \right.
     \end{array} 
\end{eqnarray*}
In encoding this into a program matrix, we also create an empty (zero) row for a rule with a head atom that does not appear in the program, so there will be \(36 = 6 \times 6\) rules for (1).

For encoding rules with \(U_{j,q}\) in the head, consider the following: (a) since we know \(U_{1,1}\) is always true (starting point), \(U_{1,q} (2 \leq q \leq 6)\) will always be false, and (b) for atoms \(U_{j,q \, (j \neq 1)}\), \(U_{j, 1}\) is always false, and for each \(q\) in \(2 \leq q \leq 6\), we generate 6 rules, i.e., there will be \(31 = 1 + (5 \times 6)\). Therefore, there will be \(161 = 6 + (5 \times 31)\) rules for (2).

By combining the program matrices for (1) and (2) (because (3) is a fact, it is omitted here), we obtain a \(197 \times 144\) program matrix \(\matr{C}_3\).

\subsection{Encoding (4) (5) (6) into constraints}

Encoding of (5) is straightforward and results in 5 constraints \((2 \leq i \leq 6)\). 
Encoding of (4) involves an XOR constraint for each vertex, for example, for vertex \(i = 1\), we have the following constraint: \(\leftarrow \neg H_{1,1} \wedge \neg H_{2,1} \wedge \cdots \wedge \neg H_{6,1}\).
Thus, encoding of (4) results in 6 constraints.
For encoding (6), we construct these constraints in two parts: (a) each vertex is visited at least once, and (b) each vertex is visited at most once.
The first part (a) is straightforward: for example, the constraint for \(i=2\) is \(\leftarrow \neg U_{2,1} \wedge \neg U_{2,2} \wedge \cdots \wedge \neg U_{2,6}\).
Thus, (a) results in 6 constraints.
The encoding of (b) requires \(\textrm{C}^5_2=10\) constraints for each \(U_j (j \geq 2)\). For example, for \(U_2\):
\begin{eqnarray*}
     \begin{array}{ll}
     \left\{\;
          \begin{array}{lll}
            \leftarrow U_{2,2} \wedge U_{2,3} \\
            \leftarrow U_{2,2} \wedge U_{2,4} \\
            \vdots \\
            \leftarrow U_{2,2} \wedge U_{2,6} \\
            \leftarrow U_{2,3} \wedge U_{2,4} \\
            \vdots \\
            \leftarrow U_{2,3} \wedge U_{2,6} \\
            \leftarrow U_{2,4} \wedge U_{2,5} \\
            \leftarrow U_{2,4} \wedge U_{2,6} \\
            \leftarrow U_{2,5} \wedge U_{2,6} \\
          \end{array}
     \right.
     \end{array} 
\end{eqnarray*}
Thus, (b) results in 50 constraints.
By combining all constraints, we obtain \(67 = 5 + 6 + 6 + 50\) constraints \(K_3\).

\end{appendix}

%Neuro-Symbolic \cite{Saker21}
%\subsection{}\label{s1.1}

%%%%%%%%%%% The bibliography starts:

%%%%%%%%%%%%%%%%%%%%%%%%%%%%%%%%%%%%%%%%%%%%%%%%%%%%%%%%%%%%%
%%                  The Bibliography                       %%
%%                                                         %%
%%  ios1.bst will be used to                               %%
%%  create a .BBL file for submission.                     %%
%%                                                         %%
%%                                                         %%
%%  Note that the displayed Bibliography will not          %%
%%  necessarily be rendered by Latex exactly as specified  %%
%%  in the online Instructions for Authors.                %%
%%                                                         %%
%%%%%%%%%%%%%%%%%%%%%%%%%%%%%%%%%%%%%%%%%%%%%%%%%%%%%%%%%%%%%

%\nocite{*}
% if your bibliography is in bibtex format, use those commands:
\bibliographystyle{ios1}           % Style BST file.
%\bibliography{../Ref}

% or include bibliography directly:
%\begin{thebibliography}{0}
%\bibitem{r1} F. Author, Information about cited object.
%
%\bibitem{r2} S. Author and T. Author, Information about cited object.
%\end{thebibliography}

\end{document}